\PassOptionsToPackage{dvipsnames}{xcolor}
\PassOptionsToPackage{expansion=false}{microtype}
\documentclass{template/bytedance}

\usepackage[utf8]{inputenc} 
\usepackage{amsfonts}       
\usepackage{nicefrac}       

\usepackage{amsmath} 
\usepackage{enumerate}
\usepackage{enumitem}
\usepackage{wrapfig}
\usepackage{graphicx}
\usepackage{rotating}
\usepackage{colortbl}
\usepackage{float}
\usepackage{placeins}
\usepackage{longtable}
\usepackage{pdflscape}
\usepackage{pgfplots}
\pgfplotsset{compat=1.16}
\usetikzlibrary{arrows.meta,positioning,shapes.geometric,calc,fadings,decorations.pathmorphing}

\fancypagestyle{firststyle}{%
  \fancyhf{}%
  \fancyhead[L]{\vskip 4mm\includegraphics[width=43.3mm,height=10mm]{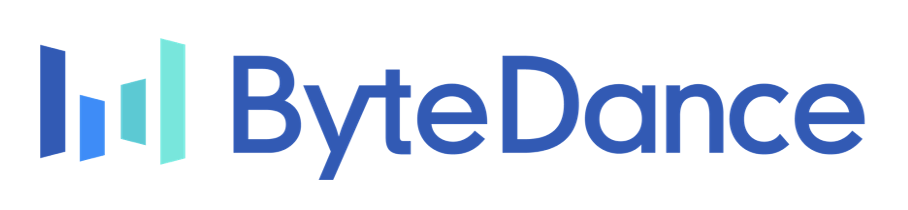}}%
}

\definecolor{autoskillblue}{RGB}{220,232,250}   
\definecolor{autoskilldeep}{RGB}{44,90,181}     
\definecolor{autoskilldark}{RGB}{24,55,130}     

\definecolor{codexamber}{RGB}{224,139,60}
\definecolor{codexamberlight}{RGB}{255,221,184}
\definecolor{codexamberdark}{RGB}{162,93,28}
\definecolor{bdteal}{RGB}{69,191,199}
\definecolor{bdteallight}{RGB}{198,233,235}
\definecolor{bdtealdark}{RGB}{36,128,134}
\definecolor{claudeyellow}{RGB}{129,88,199}
\definecolor{claudeyellowlight}{RGB}{226,216,255}
\definecolor{claudeyellowdark}{RGB}{84,49,145}
\definecolor{deltagreen}{RGB}{22,128,57}
\definecolor{skillsbenchbg}{RGB}{245,249,255}
\definecolor{skilllearnbg}{RGB}{255,248,237}

\usepackage{pifont}
\usepackage{wasysym}

\definecolor{markgreen}{RGB}{34,139,52}
\definecolor{markred}{RGB}{200,40,40}
\definecolor{markamber}{RGB}{210,150,0}
\newcommand{\cmark}{\textcolor{markgreen}{\ding{51}}}
\newcommand{\xmark}{\textcolor{markred}{\ding{55}}}
\newcommand{\pmark}{\textcolor{markamber}{$\boldsymbol{\triangle}$}}
\newcommand{\fitautotable}[1]{%
  {\footnotesize\renewcommand{\arraystretch}{0.96}%
  \resizebox{\ifdim\width>0.94\linewidth 0.94\linewidth\else\width\fi}{!}{#1}%
  }%
}

\title{MUSE-Autoskill: Self-Evolving Agents via Skill Creation, Memory, Management, and Evaluation}

\author[1,2,*]{Huawei Lin}
\author[1]{Peng Li}
\author[1]{Jie Song}
\author[1]{Fuxin Jiang}
\author[1,\dagger]{Tieying Zhang}

\affiliation[1]{ByteDance Inc.}
\affiliation[2]{Rochester Institute of Technology}

\contribution[*]{Work done during an internship at the ByteBrain team}
\contribution[\dagger]{Corresponding author}

\abstract{
Large language model (LLM) agents rely on reusable skills to solve complex tasks, but existing skill creation approaches often treat skills as isolated, static artifacts, limiting reusability, reliability, and long-term improvement. We propose \textbf{MUSE-Autoskill Agent} (\textbf{M}emory-\textbf{U}tilizing \textbf{S}kill \textbf{E}volution), a skill-centric agent framework that creates, reuses, and refines skills under a unified lifecycle: creation, memory, management, evaluation, and refinement. MUSE creates skills on demand, stores them across tasks, retrieves them through a skill catalog, and accumulates per-skill experience for later reuse and adaptation.
Across the main reported settings on SkillsBench and SkillLearnBench, MUSE-Autoskill outperforms Hermes, Codex, and Claude Code. On SkillsBench, its self-created skills surpass human-authored skills on the successfully covered subset (85.24\% vs.\ 81.17\%), showing that lifecycle-managed skills can distill agent experience into highly effective reusable assets; MUSE-created skills also transfer to Hermes more effectively than Codex- or Claude-created skills, reaching 51.90\% accuracy under transfer. These results highlight the importance of treating skills as long-lived, experience-aware, and testable~assets.
}

\date{\today}
\correspondence{Tieying Zhang, \email{tieying.zhang@bytedance.com}}

\begin{document}

\maketitle
\vspace{-.18in}

\begin{figure}[H]
\vspace{-.1in}
\centering
\begin{tikzpicture}
\begin{axis}[
    ybar=1pt,
    width=\textwidth,
    height=5.0cm,
    bar width=5pt,
    enlarge x limits=0.14,
    ymin=0, ymax=118,
    ytick={0,20,40,60,80,100},
    ylabel={Accuracy (\%)},
    ylabel style={font=\normalsize, yshift=-1mm},
    symbolic x coords={{Sci.\,\&\,Eng.}, {Data Analysis}, {Document Proc.}, {Ops\,\&\,Planning}, {SkillsBench}, {SkillLearnBench}},
    xtick=data,
    xticklabels={{Sci.\,\&\,Eng.}, {Data Analysis}, {Document Proc.}, {Ops\,\&\,Planning}, {Overall}, {Overall}},
    x tick label style={font=\small},
    y tick label style={font=\small},
    axis x line*=bottom,
    axis y line*=left,
    axis line style={draw=gray!55},
    tick style={draw=gray!55},
    area legend,
    legend image code/.code={\draw [#1] (0cm,-0.08cm) rectangle (0.32cm,0.12cm);},
    legend style={
        font=\footnotesize,
        at={(0.5,1.08)}, anchor=south, legend columns=4,
        draw=gray!40, fill=white, rounded corners=2pt,
        /tikz/every even column/.append style={column sep=8pt},
    },
    legend cell align={left},
    set layers,
    ymajorgrids=false,
    xmajorgrids=false,
    major grid style={black!55, dash pattern=on 4pt off 2pt, line width=0.8pt},
    axis on top=true,
    clip mode=individual,
    nodes near coords,
    every node near coord/.append style={
        font=\fontsize{8}{8}\selectfont,
        rotate=90,
        anchor=west,
        inner xsep=0pt, inner ysep=1pt,
        xshift=2pt,
        yshift=0pt,
        /pgf/number format/.cd, fixed, fixed zerofill=false, precision=1,
    },
]
\path[draw=none, fill=skillsbenchbg, fill opacity=0.45, on layer=axis background]
  (rel axis cs:0,0) rectangle (rel axis cs:0.8125,1);
\path[draw=none, fill=skilllearnbg, fill opacity=0.45, on layer=axis background]
  (rel axis cs:0.8125,0) rectangle (rel axis cs:1,1);
\draw[markred!55, dashed, line width=0.7pt]
  (rel axis cs:0.8125,0) -- (rel axis cs:0.8125,1);
\node[
    font=\footnotesize\bfseries,
    text=autoskilldark,
    inner xsep=2pt, inner ysep=1pt
] at (rel axis cs:0.41,0.925) {SkillsBench};
\node[
    font=\footnotesize\bfseries,
    text=autoskilldark,
    inner xsep=2pt, inner ysep=1pt
] at (rel axis cs:0.91,0.925) {SkillLearnBench};
\draw[gray!35, dash pattern=on 3pt off 2pt, line width=0.25pt]
  (rel axis cs:0,0.1695) -- (rel axis cs:1,0.1695);
\draw[gray!35, dash pattern=on 3pt off 2pt, line width=0.25pt]
  (rel axis cs:0,0.3390) -- (rel axis cs:1,0.3390);
\draw[gray!35, dash pattern=on 3pt off 2pt, line width=0.25pt]
  (rel axis cs:0,0.5085) -- (rel axis cs:1,0.5085);
\draw[gray!35, dash pattern=on 3pt off 2pt, line width=0.25pt]
  (rel axis cs:0,0.6780) -- (rel axis cs:1,0.6780);
\draw[gray!35, dash pattern=on 3pt off 2pt, line width=0.25pt]
  (rel axis cs:0,0.8475) -- (rel axis cs:1,0.8475);
\draw[gray!55, line width=0.4pt]
  (rel axis cs:0,1) -- (rel axis cs:1,1) -- (rel axis cs:1,0);
\addplot[draw=bdtealdark, fill=bdteallight]
  coordinates {({Sci.\,\&\,Eng.},40.61) ({Data Analysis},36.34) ({Document Proc.},54.29) ({Ops\,\&\,Planning},25.92) ({SkillsBench},37.24) ({SkillLearnBench},37.0)};
\addplot[draw=bdtealdark, fill=bdteal]
  coordinates {({Sci.\,\&\,Eng.},58.89) ({Data Analysis},40.60) ({Document Proc.},58.57) ({Ops\,\&\,Planning},39.64) ({SkillsBench},48.02) ({SkillLearnBench},70.0)};
\addplot[draw=codexamberdark, fill=codexamberlight]
  coordinates {({Sci.\,\&\,Eng.},54.52) ({Data Analysis},38.31) ({Document Proc.},68.57) ({Ops\,\&\,Planning},29.16) ({SkillsBench},44.80) ({SkillLearnBench},39.0)};
\addplot[draw=codexamberdark, fill=codexamber]
  coordinates {({Sci.\,\&\,Eng.},67.12) ({Data Analysis},50.18) ({Document Proc.},72.86) ({Ops\,\&\,Planning},47.48) ({SkillsBench},57.58) ({SkillLearnBench},68.0)};
\addplot[draw=claudeyellowdark, fill=claudeyellowlight]
  coordinates {({Sci.\,\&\,Eng.},52.33) ({Data Analysis},38.05) ({Document Proc.},62.86) ({Ops\,\&\,Planning},27.00) ({SkillsBench},42.43) ({SkillLearnBench},33.0)};
\addplot[draw=claudeyellowdark, fill=claudeyellow]
  coordinates {({Sci.\,\&\,Eng.},63.88) ({Data Analysis},52.59) ({Document Proc.},64.29) ({Ops\,\&\,Planning},48.60) ({SkillsBench},56.15) ({SkillLearnBench},63.0)};
\addplot[draw=autoskilldark, fill=autoskillblue]
  coordinates {({Sci.\,\&\,Eng.},54.57) ({Data Analysis},39.49) ({Document Proc.},67.14) ({Ops\,\&\,Planning},35.52) ({SkillsBench},46.95) ({SkillLearnBench},43.0)};
\addplot[draw=autoskilldark, fill=autoskilldeep]
  coordinates {({Sci.\,\&\,Eng.},67.97) ({Data Analysis},51.48) ({Document Proc.},74.29) ({Ops\,\&\,Planning},51.40) ({SkillsBench},59.67) ({SkillLearnBench},72.0)};
\legend{Hermes w/o, Hermes w/ hum, Codex w/o, Codex w/ hum, Claude Code w/o, Claude Code w/ hum, \textbf{MUSE w/o (Ours)}, \textbf{MUSE w/ hum (Ours)}}
\end{axis}
\end{tikzpicture}
\vspace{-.1in}
\caption{\textbf{MUSE-Autoskill (ours) leads across SkillsBench and SkillLearnBench.}
Accuracy (\%) of four GPT-5.5-backed agents. The first four groups are the 75-task SkillsBench super-domains; \textit{SkillsBench} is the strict all-task average; \textit{SkillLearnBench} is the 100-instance benchmark average. Paired bars per agent: lighter = \textit{without skills}, saturated = \textit{with human skills}. MUSE-Autoskill achieves the highest human-skill score in 3 of 4 SkillsBench domains, on the SkillsBench average (59.7\%), and on SkillLearnBench (72.0\%). See Section~\ref{sec:experiments} and Tables~\ref{tab:skill_effect}--\ref{tab:per_domain}.}
\label{fig:per_domain}
\end{figure}
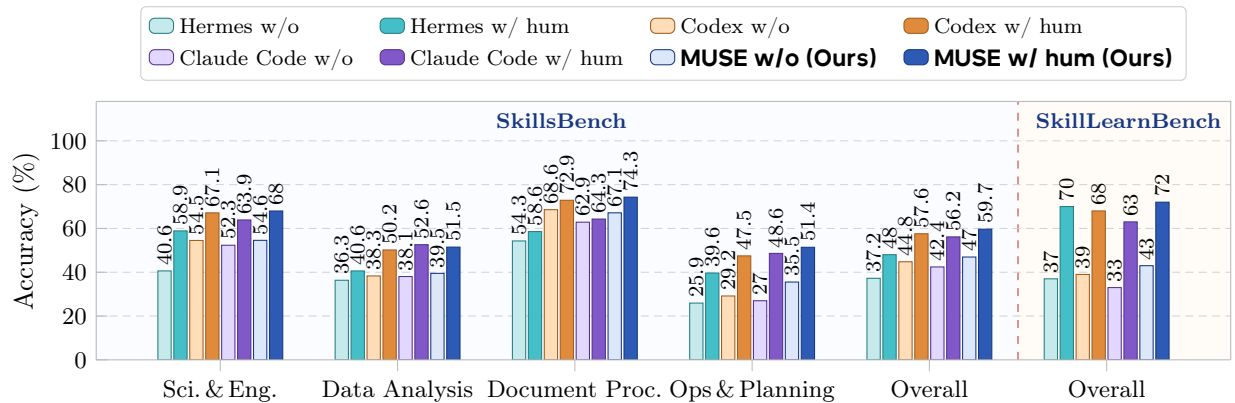

\section{Introduction}

\paragraph{Skills for agents.}
Large language model (LLM) agents are increasingly tasked with solving complex, real-world problems that involve interacting with external tools, data, and code, often spanning many steps and disparate domains~\cite{react, gaia, swebench}. As task scope grows, raw model reasoning alone is insufficient: agents need access to reusable units of capability, namely \emph{skills}, that encapsulate procedures, executable code, or domain-specific instructions and can be composed into solutions~\cite{voyager, anthropic-skills}. Skills are emerging as the natural abstraction for scalable agent systems because they decouple capability from monolithic model weights, enabling modular execution and the accumulation of structured domain knowledge~\cite{anthropic-skills, agentskills-survey}. The central open question is how to \emph{enable agents to continuously improve their capabilities} through skills they can obtain, organize, and refine on their own, without relying on human authoring at every step.

\paragraph{Limits of AutoSkill.}
A growing line of work uses LLMs to synthesize skills automatically, starting from Voyager's executable code library in Minecraft~\cite{voyager} and extending to general-purpose agents via AutoSkill~\cite{yang2026autoskill}, EvoSkill~\cite{alzubi2026evoskill}, and SkillGen~\cite{skillgen}. More recent approaches use reinforcement learning to jointly optimize skill selection, use, and distillation (Skill1~\cite{skill1}) or to train a dedicated skill curator (SkillOS~\cite{skillos}). On the production side, Anthropic's Agent Skills~\cite{anthropic-skills} standardize skills as portable folders of instructions and scripts. While these methods successfully expand agent functionality, they typically cover only part of the skill lifecycle and leave four practical gaps: (i) a \emph{creation--usage mismatch}, where skills are produced without access to the agent's runtime context; (ii) \emph{no structured per-skill memory} that accumulates free-form experience about individual skills across tasks; (iii) \emph{static, unvalidated skills} without unit-test-driven evaluation or refinement; and (iv) \emph{poor context handling}, where flat conversation histories truncate or overflow on long-horizon tasks.

\paragraph{Skill lifecycle.}
We argue that skills should not be one-off generation outputs but \emph{long-lived, evolving assets} of an agent system. A useful skill is created on demand within the agent's reasoning loop, stored with associated experience and metadata~\cite{memgpt, generativeagents, reflexion}, retrieved when contextually relevant, validated through tests and runtime feedback, and continuously refined as new evidence accumulates~\cite{selfdebug, selfrefine, skillgen}. We formalize this perspective as a unified \emph{skill lifecycle} with five stages: \textbf{creation, memory, management, evaluation, and refinement}. This reframing turns skills from disposable artifacts into managed, testable, and transferable infrastructure: the foundation needed for agents to accumulate experience across tasks, sessions, and even across different agent systems.

\begin{figure}[t]
    \hspace{-.1in}
    \includegraphics[width=1.02\linewidth]{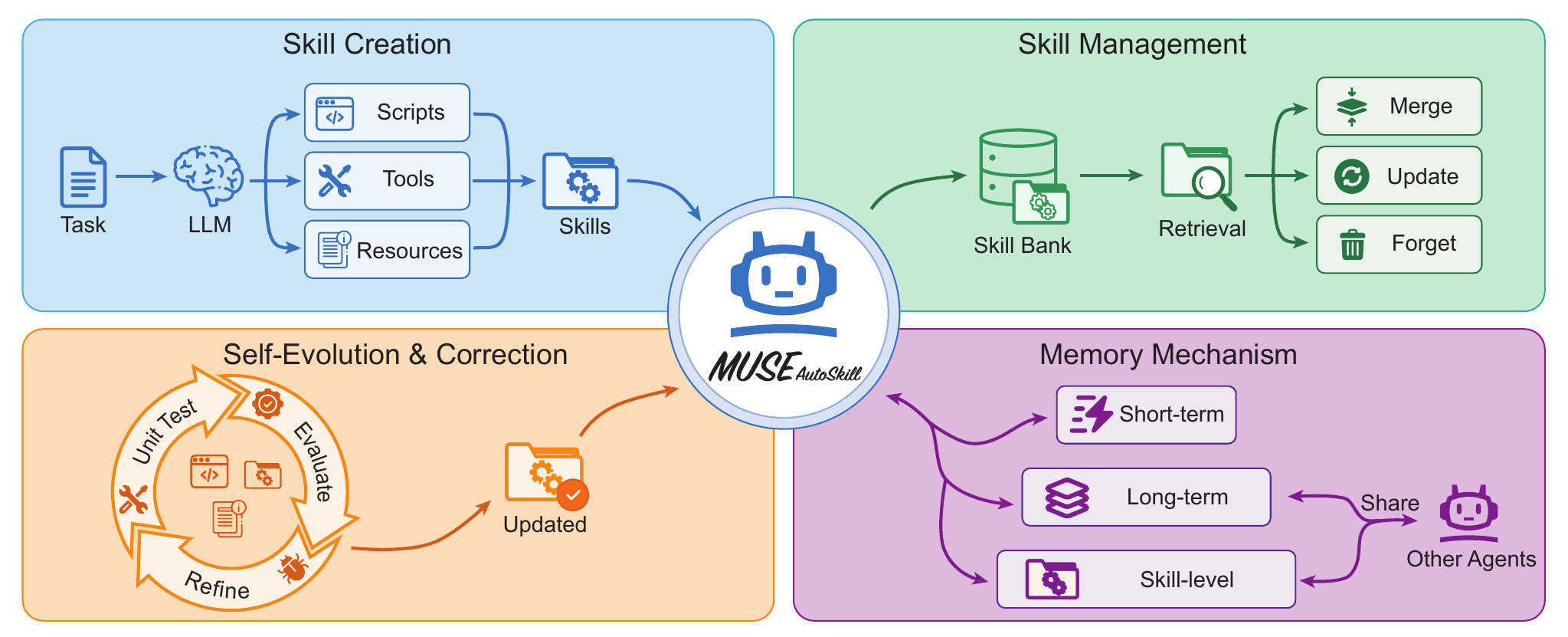}
\vspace{-.2in}
    \caption{\textbf{MUSE-Autoskill Agent architecture.} MUSE organizes skills into a unified lifecycle of creation, memory, management, evaluation, and refinement, enabling agents to generate, refine, and reuse skills with accumulated experience over time.}
    \label{fig:overview}
  \vspace{-.05in}
\end{figure}

\paragraph{MUSE-Autoskill framework.}
We instantiate this lifecycle in \textbf{MUSE-Autoskill Agent} (\textbf{M}emory-\textbf{U}tilizing \textbf{S}kill \textbf{E}volution; Figure~\ref{fig:overview}). MUSE tightly couples skill creation with execution through a built-in \texttt{skill\_create} tool invoked from within the runtime loop, eliminating the creation--usage mismatch. It introduces a \emph{multi-level memory} comprising short-term, long-term, and (uniquely) \emph{skill-level memory}, which accumulates per-skill experience across tasks and informs future invocations. An evaluation subsystem grounds reliability in unit tests and execution feedback, automatically triggering refinement when tests fail. A structured context manager with adaptive compression and cross-session state persistence supports long-horizon tasks without information loss or context-window blowup. Together, these components make skills externalized, testable, and transferable, rather than internal model behavior locked inside opaque weights.

\paragraph{Results.}
Figure~\ref{fig:per_domain} previews our headline results on \textbf{SkillsBench}, a benchmark of real-world tasks graded by automated verifiers in standardized Docker environments. On the 75-task common set, MUSE-Autoskill achieves the best no-skill accuracy (\textbf{46.95\%}) and the best human-skill accuracy (\textbf{59.67\%}, a $+12.72$ pp lift). Human skills improve all four GPT-5.5-backed agents by $+10.78$ to $+13.73$ pp. Self-created skills also improve every skill-creating agent on the same all-task denominator: MUSE-Autoskill reaches \textbf{53.42\%}, above Codex (47.52\%) and Claude Code (44.27\%). A stability analysis in Appendix~\ref{app:stability} further shows that MUSE-Autoskill-created skills have the lowest run-to-run reward dispersion among the compared self-created-skill conditions. Generated skills transfer across agent runtimes: Hermes reaches \textbf{51.90\%} with MUSE-Autoskill-created skills, exceeding Hermes with human skills at 48.02\%. We further evaluate on \textbf{SkillLearnBench}, a separate 20-task, 100-instance benchmark for continual skill generation, where MUSE-Autoskill also obtains the best human-skill accuracy (72.0\%) and self-created-skill accuracy (48.0\%).

\textbf{Contributions.} This paper makes four contributions:
\begin{itemize}[itemsep=2pt, topsep=2pt, leftmargin=*]
\item \textbf{Skill lifecycle.} We reframe skills from one-off generation outputs into long-lived, lifecycle-managed assets, identifying five stages (creation, memory, management, evaluation, refinement) that any practical skill-centric agent system must address.
\item \textbf{MUSE-Autoskill.} A skill-centric agent that improves its task-solving capability over time by integrating skill creation with runtime execution, evaluating code-backed skills with unit tests and using sandbox/runtime feedback for all generated skills.
\item \textbf{Infrastructure.} Multi-level memory with a novel \emph{skill-level} memory that accumulates per-skill experience across tasks; adaptive context compression with cross-session state persistence; and cross-agent skill transfer that makes generated skills usable beyond their authoring agent.
\item \textbf{Validation.} Best-in-class SkillsBench accuracy among four GPT-5.5-backed agents on the 75-task common set (59.67\% with human skills, $+12.72$ pp lift); strongest self-created skill result under all-task scoring (53.42\%); generated skills that transfer to Hermes above its human-skill score; and corroborating SkillLearnBench results on 100 verified instances.
\end{itemize}

\section{Related Work}

\subsection{LLM Agents}

LLM-based agents that interact with tools, environments, and data have advanced rapidly in recent years~\cite{DBLP:journals/corr/abs-2402-02716, DBLP:conf/emnlp/SchmidgallSWSWYLMLB25, DBLP:conf/iclr/HongZCZCWZWYLZR24, autogen}. Building on ReAct~\cite{react}'s interleaving of reasoning and action, follow-up systems extend the paradigm to broader workflows, including multimodal autonomous agents such as Agent-Omni~\cite{agent-omni}, MUSE~\cite{lu2026muse}, and OmniGAIA~\cite{li2026omnigaia}, and a wider body of work on self-improving agents~\cite{reflexion, selfrefine}. A parallel line of work focuses on equipping agents with tool-use capabilities, ranging from few-shot tool calling~\cite{DBLP:conf/nips/SchickDDRLHZCS23} to tool orchestration via model selection~\cite{DBLP:journals/corr/abs-2303-17580} and large-scale API retrieval~\cite{DBLP:conf/nips/PatilZ0G24}; for software engineering specifically, agents such as CodeAgent~\cite{DBLP:conf/acl/ZhangLLSJ24}, SWE-Agent~\cite{sweagent}, and OpenHands~\cite{openhands} drive tool-integrated workflows over sandboxed shells and editors to resolve real-world repository tasks. The capabilities of such systems are now measured by general agent benchmarks including GAIA~\cite{gaia}, SWE-bench~\cite{swebench}, and AgentBench~\cite{agentbench}, which together cover web browsing, real-world software engineering, and multi-environment tool use. Despite this progress, most agent frameworks treat the set of available actions as either a fixed, hand-engineered tool registry or a flat conversational scratchpad. They do not natively support agents that can \emph{author, validate, and accumulate} their own reusable capabilities over time, which is precisely the gap the skill-centric literature, and our framework, set out to close.

\subsection{Automatic Skill Systems}

We organize the growing literature on automatic skill systems along two axes: which stages of the skill lifecycle (\emph{creation, memory, management, evaluation, refinement}) a method addresses, and whether it operates entirely at inference time or requires additional model training. Table~\ref{tab:related_work} summarizes the resulting comparison along these two axes.

The first major direction builds skill systems on top of pretrained LLMs without any fine-tuning. Voyager~\cite{voyager} is the seminal example: in the Minecraft setting, it maintains an ever-growing library of executable-code skills, with self-verification and iterative prompting that lets the same LLM both author and refine skills in response to environment feedback. Follow-up work generalizes this paradigm to general-purpose agents: AutoSkill~\cite{yang2026autoskill} derives, maintains, and reuses skills from dialogue and interaction traces as a model-agnostic plugin layer; EvoSkill~\cite{alzubi2026evoskill} analyses execution failures and proposes new skills or edits, retaining only those that improve held-out validation under a Pareto-frontier selection; and SkillGen~\cite{skillgen} iteratively refines skills via contrastive induction over successful and failed trajectories, modelling each skill as an intervention to empirically verify its net effect. The feedback-driven refinement underlying these methods is rooted in a broader self-improvement literature outside the skill setting: Reflexion~\cite{reflexion} maintains reflective text in an episodic memory buffer across attempts, Self-Refine~\cite{selfrefine} iteratively rewrites outputs using self-generated critiques, Self-Debug~\cite{selfdebug} closes the loop on code generation with execution and unit-test traces, and ExpeL~\cite{expel} extracts natural-language insights across training tasks for inference-time reuse. These methods all improve agent behavior through linguistic feedback but stop short of treating skills as first-class, externalized, testable artifacts that outlive a single task or agent. On the industrial side, Anthropic's Agent Skills~\cite{anthropic-skills} standardize skills as portable folders of \texttt{SKILL.md} instructions and scripts loaded via progressive disclosure; this is the closest practical analogue of our externalized skill format, but the system leaves evaluation and refinement to human authoring. Collectively, these training-free methods are lightweight and naturally portable across LLM backbones, yet each covers only part of the lifecycle: none simultaneously supports structured per-skill memory, unit-test-driven evaluation, and automatic refinement triggered by test feedback.

A second, concurrent direction uses reinforcement learning to optimize skill behavior jointly with the policy. SkillMaster~\cite{skillmaster} learns a single policy that both acts and edits its skill bank, with edits credited by counterfactual downstream utility. Skill1~\cite{skill1} frames skill evolution as a unified RL problem, co-optimizing skill selection, utilization, and distillation under a shared task-outcome reward. SkillOS~\cite{skillos} pairs a frozen executor with a trainable curator that updates an external skill repository from accumulated experience, and shows that the curator generalizes across executor backbones; this is a portability axis complementary to ours, where the skills themselves rather than the curator are the unit of transfer. Youtu-Agent~\cite{youtuagent} pursues a related direction via hybrid policy optimization of tools and agent configurations. These RL-based methods can attain strong optimality on the environments they are trained on, but they couple skill behavior to a trained policy or curator: migrating to a new backbone typically requires additional training, and skills produced by one trained policy are not directly usable by a different agent without re-training.

\begin{table}[t]
\caption{Related work on automatic skill systems by lifecycle stage. \cmark{} = covered; \pmark{} = partial; \xmark{} = not addressed. \emph{Memory} = persistent per-skill experience across tasks. \emph{Cross-agent} = skills from one agent are usable by another without modification; \cmark{} requires an explicit cross-agent transfer experiment, \pmark{} indicates portability only across LLM backbones or product variants of the same agent. \emph{Training-free} = inference-time only, no fine-tuning or RL.}
  \centering
  \small
  \setlength{\tabcolsep}{4pt}
  \fitautotable{%
  \begin{tabular}{lccccccc}
  \toprule
  & \multicolumn{5}{c}{\textbf{Lifecycle stage}} & & \\
  \cmidrule(lr){2-6}
  \textbf{Method} & \textbf{Creation} & \textbf{Memory} & \textbf{Management} & \textbf{Evaluation} & \textbf{Refinement} & \textbf{Cross-agent} & \textbf{Training-free} \\
  \midrule
  Voyager~\cite{voyager}                          & \cmark & \xmark & \cmark & \pmark & \cmark & \pmark & \cmark \\
  AutoSkill~\cite{yang2026autoskill}              & \cmark & \pmark & \pmark & \xmark & \cmark & \xmark & \cmark \\
  EvoSkill~\cite{alzubi2026evoskill}              & \cmark & \xmark & \pmark & \cmark & \cmark & \xmark & \cmark \\
  SkillGen~\cite{skillgen}                         & \cmark & \xmark & \xmark & \cmark & \cmark & \pmark & \cmark \\
  Anthropic Skills~\cite{anthropic-skills}        & \cmark & \xmark & \cmark & \xmark & \xmark & \pmark & \cmark \\
  SkillMaster~\cite{skillmaster}                   & \cmark & \xmark & \pmark & \pmark & \cmark & \xmark & \xmark \\
  Youtu-Agent~\cite{youtuagent}                    & \cmark & \xmark & \pmark & \xmark & \xmark & \xmark & \xmark \\
  Skill1~\cite{skill1}                             & \cmark & \pmark & \pmark & \pmark & \pmark & \xmark & \xmark \\
  SkillOS~\cite{skillos}                           & \cmark & \cmark & \cmark & \pmark & \cmark & \xmark & \xmark \\
  \midrule
  \rowcolor{autoskillblue}
  \textbf{MUSE-Autoskill (Ours)}                   & \cmark & \cmark & \cmark & \cmark & \cmark & \cmark & \cmark \\
  \bottomrule
  \end{tabular}%
  }
  
  \label{tab:related_work}
  
\end{table}

\subsection{Benchmarks and Positioning}

Several recent benchmarks complement the methods above by stressing different lifecycle stages. SkillsBench~\cite{skillsbench}, which we adopt in our experiments, measures end-to-end task accuracy with and without skills across diverse Docker-evaluated real-world tasks. SkillRet~\cite{skillret} isolates the management stage by evaluating skill retrieval at scale from a library of nearly 18,000 community-contributed skills. SkillLearnBench~\cite{skilllearnbench} and LifelongAgentBench~\cite{lifelongagentbench} focus on continual and lifelong skill acquisition over task streams, and notably report that strong individual methods do not consistently dominate, motivating system-level designs such as ours. A concurrent survey~\cite{agentskills-survey} catalogues skill-acquisition modalities and architectural choices for LLM agents, situating both training-free and training-based directions within a broader taxonomy.

Compared with the methods above, MUSE-Autoskill differs in that it brings all five lifecycle stages together within a single training-free framework, rather than addressing creation or refinement in isolation. In particular, it introduces skill-level memory that accumulates per-skill experience across tasks, uses unit-test-driven evaluation that automatically triggers refinement when tests fail, and is the only general-purpose method to \emph{empirically validate} cross-agent skill transfer by injecting its generated skills into a different agent without modification (Section~\ref{sec:experiments}); other portability claims in the literature are limited to swapping the underlying LLM backbone or sharing skills across product variants of the same agent family, without an explicit cross-agent experiment. The combination of full lifecycle coverage and a training-free design also makes the system portable across LLMs and agent architectures, as summarized in the bottom row of Table~\ref{tab:related_work}.

\begin{figure}[t]
  \centering
  \vspace{-.1in}
  \includegraphics[width=\textwidth]{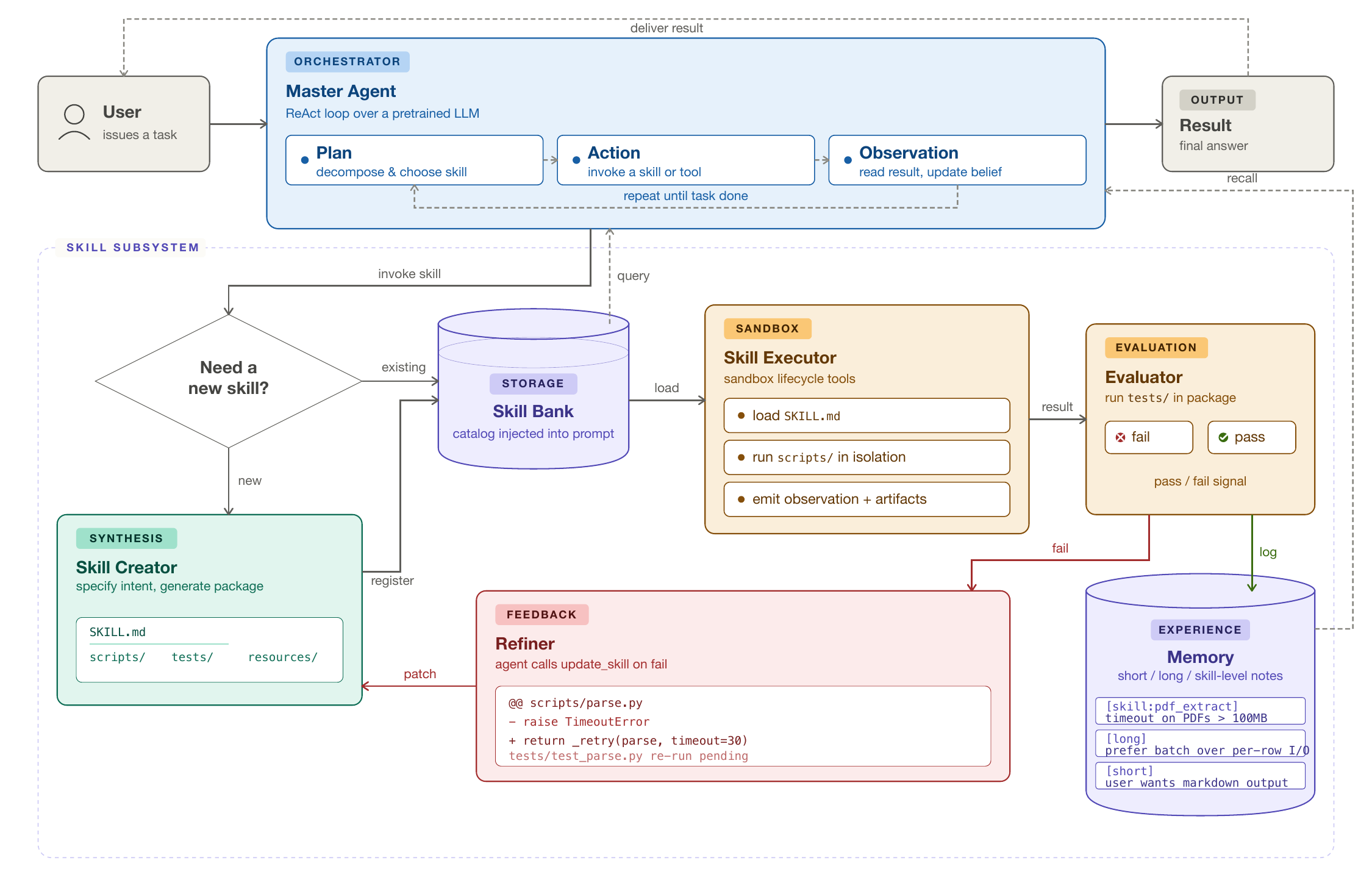}
\vspace{-.2in}
  \caption{\textbf{End-to-end flow of MUSE-Autoskill.} The Master Agent runs a ReAct loop; when a skill is needed it either retrieves one from the \emph{Skill Bank} or dispatches the \emph{Skill Creator} to synthesize a new package (\texttt{SKILL.md} plus optional \texttt{scripts/} and, for code-backed skills, \texttt{tests/}). The \emph{Evaluator} runs bundled tests when present and otherwise relies on sandbox/runtime checks; failed checks send the package to the \emph{Refiner}, while accepted observations are appended to \emph{Memory} and surfaced on later steps.}
  \label{fig:agent_flow}
  \vspace{-.15in}
  \end{figure}

\section{MUSE-Autoskill Agent}

In this section, we present \textbf{MUSE-Autoskill Agent}, a skill-centric agent framework that solves complex tasks by dynamically creating, reusing, and refining skills. MUSE integrates skill creation, execution, memory, management, and evaluation within a unified agent loop. Figure~\ref{fig:overview} illustrates the overall architecture and the five lifecycle stages described below.

\subsection{Agent Framework}

The agent operates in an iterative decision-making loop consisting of three core stages: \textit{Planning}, \textit{Action}, and \textit{Observation}~\cite{react}. Given an input query, the agent continuously cycles through these stages to progressively solve the task. This design enables dynamic reasoning, skill invocation, and adaptive refinement based on intermediate feedback from tool calls and skill executions.

\paragraph{Planning}
In the planning stage, the agent interprets the input query and determines the next step toward achieving the task objective. This involves decomposing the problem, selecting appropriate strategies, and deciding whether to invoke external skills. The agent may also leverage past observations and memory to refine its plan, enabling more informed and context-aware decisions.

\paragraph{Action}
In the action stage, the agent executes the planned step by invoking skills. These may include retrieving existing skills from the skill bank or utilizing built-in functions such as skill creation and web search. The selected skill is invoked within the agent's ReAct loop using its built-in tools, producing intermediate or final outputs for the task. The detailed execution mechanism of skills will be introduced in Section~\ref{sec:skill_execution}.

\paragraph{Observation}
In the observation stage, the agent collects and analyzes the results returned from execution. These observations are used to evaluate progress toward the goal and to inform subsequent planning decisions. Through this feedback loop, the agent can iteratively refine its behavior, handle errors, and adapt to complex, multi-step tasks in practice.

\subsection{Skill Lifecycle}
\label{sec:skill_lifecycle}

As illustrated in Figure~\ref{fig:agent_flow}, the agent organizes skills into a unified lifecycle of five stages: \textit{creation}, \textit{memory}, \textit{management}, \textit{evaluation}, and \textit{refinement}. To bootstrap this process, the agent is equipped with a small set of built-in skills, including \textit{skill\_create} and \textit{web\_search}. In the autonomous self-creation setting, new task-specific skills are produced through this mechanism; in evaluation settings, the same runtime can also load externally supplied human skills through the skill bank.

\paragraph{Skill}
As illustrated in Figure~\ref{fig:overview}, a \textit{skill} is the basic unit of execution in our system. Each skill is packaged as a structured directory with standard components, following Anthropic's Agent Skills format~\cite{anthropic-skills}. It includes a \texttt{SKILL.md} file that defines its interface, such as name, description, inputs, and outputs, and may also include subdirectories like \texttt{scripts/} for executable code, \texttt{resources/} for auxiliary data, and, when the skill includes generated code, \texttt{tests/} for validation.

Skills are executed through a unified interface. At runtime, the agent reads \texttt{SKILL.md} to understand how to use the skill, and decides whether to read resources, run scripts, or both. If scripts are required, the execution engine runs the corresponding code with the given inputs and returns the outputs.

Using skills changes repeated work from open-ended reasoning into a shorter procedure call. The agent can load only the skill interface first, then read the full body or run bundled scripts only when needed; reused skills therefore reduce repeated exploration and make later runs more direct.

\paragraph{Skill Creation}
As illustrated in Figure~\ref{fig:overview}, new skills are generated through the built-in \textit{skill\_create} skill. When existing skills are not sufficient, the agent provides a high-level specification of the desired functionality, including its purpose, inputs, and expected outputs.

Based on this specification, the system follows a structured pipeline to construct the skill. It first generates the \texttt{SKILL.md} file to define the interface, then plans the internal structure such as \texttt{scripts/}, \texttt{resources/}, and, for code-backed skills, \texttt{tests/}, and finally generates the corresponding files. The result is a complete and executable skill package.

After creation, the skill is checked before registration. Code-backed skills may include a \texttt{tests/} directory; when present, the system runs those tests inside the sandbox and blocks registration until they pass. For text-only procedural skills and code-backed packages without generated tests, the system falls back to sandbox execution checks and runtime feedback from the source trajectory. If a check fails, the agent inspects the error trace and invokes \texttt{update\_skill} to patch the package before rechecking it. This create $\rightarrow$ evaluate $\rightarrow$ register loop keeps generated skills inspectable and gives the agent a concrete failure signal for later refinement.

\paragraph{Skill Evaluation}
As illustrated in Figure~\ref{fig:overview}, skills are evaluated before they are reused. For code-backed skills, the strongest signal comes from unit tests when the package contains a \texttt{tests/} directory. For text-only procedural skills, and for code-backed skills without generated tests, the evaluator uses sandbox execution, source-trajectory checks, and runtime feedback as weaker but still useful validation signals.

This process filters out many incorrect or unstable packages and records the reason for failure. As part of the self-evolution loop shown in Figure~\ref{fig:overview}, failed tests or failed execution checks can trigger updates or regeneration of the skill. The result is not a guarantee of correctness, but a concrete audit path: each accepted skill has either passed generated tests or survived the available sandbox/runtime checks.

\paragraph{Skill Execution}\label{sec:skill_execution}
As illustrated in Figure~\ref{fig:overview}, skill execution is carried out within the agent's ReAct loop using its built-in tools. Given a task, the agent reads the available skill catalog and selects an appropriate skill. It then reads the \texttt{SKILL.md} file to understand the skill interface, standard operating procedure, and required components.

Following the procedure defined in \texttt{SKILL.md}, the agent decides whether to read from \texttt{resources/}, execute code in \texttt{scripts/} via sandbox tools, or combine both. Code execution is mediated by a small set of sandbox lifecycle tools (\texttt{create\_sandbox}, \texttt{sandbox\_run}, \texttt{sandbox\_upload}/\texttt{sandbox\_download}, and \texttt{close\_sandbox}) that the agent invokes from inside its ReAct loop. Each sandbox is an isolated process / container with its own filesystem, so failures, side effects, and resource usage are contained per skill invocation. Rather than introducing a separate execution engine, skill execution reuses the same general-purpose tools the agent already uses (file reading, terminal commands, sandbox calls), which avoids redundant infrastructure and lets execution benefit from the agent's full reasoning capability.

The execution process is iterative: intermediate results are fed back into the agent's reasoning loop, enabling progressive refinement and error handling. This unified approach ensures consistent execution across all skills while preserving flexibility for both simple and complex tasks.

\paragraph{Skill Memory}
As illustrated in Figure~\ref{fig:overview}, the agent maintains memory at multiple levels to support skill reuse and accumulation over time. In particular, skill-level memory stores the skills themselves along with their metadata, such as descriptions, inputs, and usage history. This allows the agent to efficiently retrieve relevant skills for new tasks.

In addition, the agent appends notes and observations to short-term and long-term memory, providing context for future decisions. These notes help the agent avoid recreating a skill it has already used, remember file-format or environment quirks, and choose a known procedure before starting from scratch.

\paragraph{Skill Management}
As illustrated in Figure~\ref{fig:overview}, skill management maintains the quality and usability of the skill bank. Each skill is indexed using metadata from \texttt{SKILL.md}, including its name, description, inputs, and outputs. At the start of each task, the agent is provided with a catalog of available skills injected into the system prompt, following the progressive-disclosure pattern of Anthropic's Agent Skills~\cite{anthropic-skills}. The agent then selects the most relevant skill during planning based on this catalog, enabling efficient reuse and reducing unnecessary skill creation.

In addition to retrieval, the system supports maintenance of the skill bank through three mechanisms: \textit{refinement}, \textit{merging}, and \textit{pruning}. When a skill fails generated tests or produces incorrect outputs during execution, the agent revises or regenerates it based on the error feedback. When newly created skills overlap significantly with existing ones, the agent can merge them into a single, more general skill to avoid redundancy. Skills that consistently fail or remain unused over time can be removed from the active catalog. These operations keep the catalog smaller and make skill selection less dependent on scanning redundant entries.

\subsection{Memory}

Memory plays a central role in enabling MUSE to accumulate knowledge and reuse previously acquired capabilities. Our design builds on prior hierarchical memory architectures for LLM agents: MemGPT~\cite{memgpt} pages between in-context and external memory in an OS-style hierarchy, Generative Agents~\cite{generativeagents} maintain a memory stream with periodic synthesis into higher-level reflections, and Reflexion~\cite{reflexion} and ExpeL~\cite{expel} accumulate natural-language reflections and insights across episodes. MUSE extends these by adding a per-skill memory scope tied to each \texttt{SKILL.md} file, complementing short- and long-term layers shared with prior work.

\paragraph{Skill-level Memory} Each skill in the bank carries its own \texttt{.memory.md} file, into which the agent appends notes, lessons, and usage observations accumulated across tasks (e.g., known failure modes, input format quirks, performance caveats). When the same skill is loaded later, this per-skill memory is surfaced alongside its \texttt{SKILL.md} interface, letting the agent benefit from previously learned experience without re-deriving it.

\paragraph{Short-term Memory} Short-term memory maintains the current task context, including intermediate reasoning steps, observations, and temporary execution results. As the context grows, it is adaptively compressed by summarizing intermediate steps, allowing the agent to handle long-horizon tasks without exceeding the model's token budget during extended runs.

\paragraph{Long-term Memory} Long-term memory stores persistent notes the agent appends across sessions, including reusable conclusions, environment quirks, and general lessons learned outside any single skill (e.g., ``prefer batched I/O,'' ``the project uses pinned package versions''). Unlike short-term memory, long-term memory is not subject to compression and serves as a growing repository of accumulated experience, enabling the agent to improve decision-making over time by drawing on lessons learned in prior runs.

\subsection{Context Management}
\label{sec:context}

\begin{figure}[t]
\centering
\includegraphics[width=\textwidth]{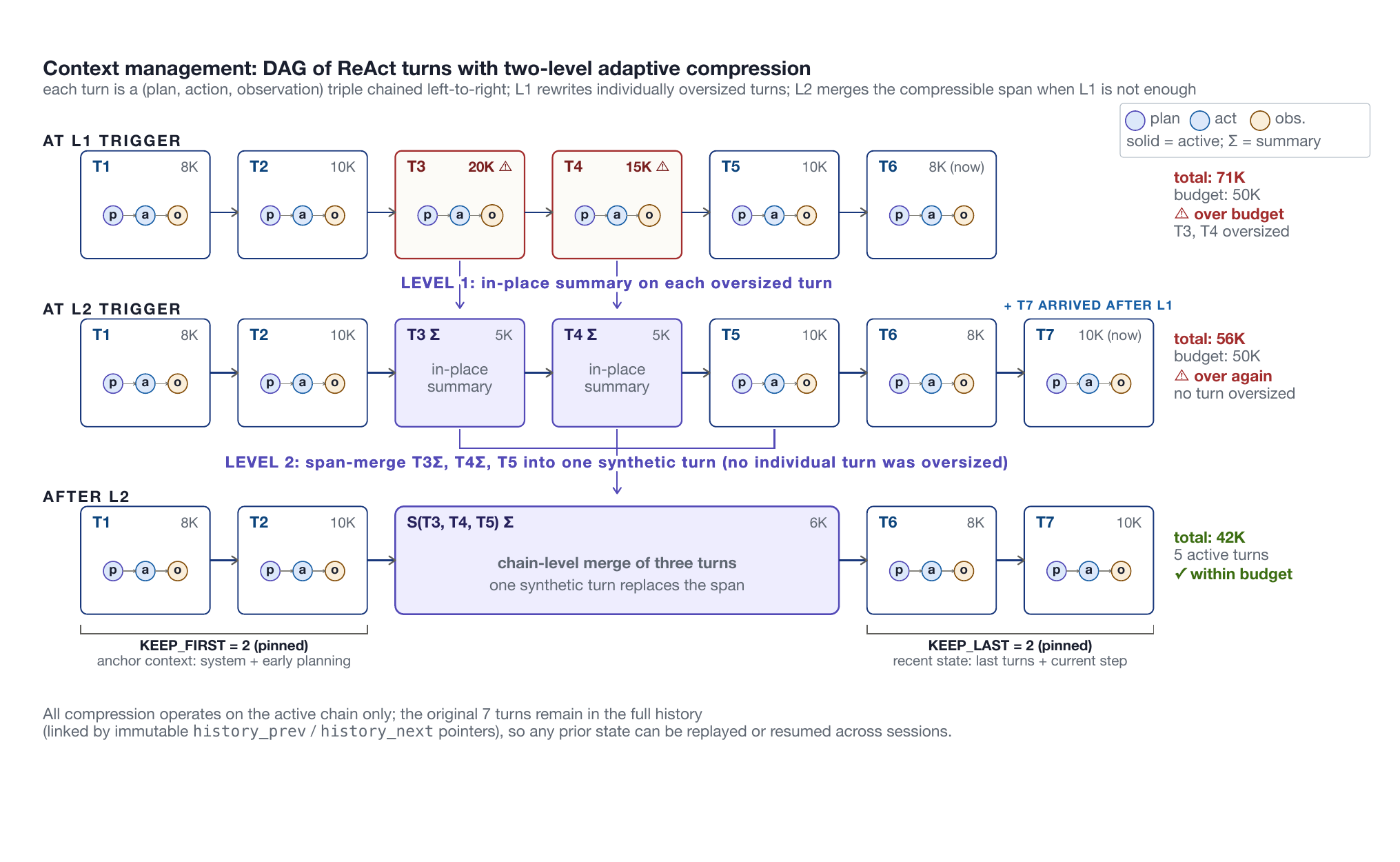}
\vspace{-.6in}
\caption{\textbf{Adaptive context compression over a DAG of ReAct turns.} Each turn is a (\emph{plan}, \emph{action}, \emph{observation}) triple; the first \texttt{KEEP\_FIRST} and last \texttt{KEEP\_LAST} turns are always pinned and only the middle is eligible for compression. \textbf{Top$\to$Middle:} Level-1 rewrites individually oversized turns in place. \textbf{Middle$\to$Bottom:} when no single turn is oversized but the chain is still over budget, Level-2 merges the compressible span into one synthetic node. Original turns remain in the full history (linked by immutable \texttt{history\_prev}/\texttt{history\_next} pointers), so the trajectory is fully replayable.}
\label{fig:context_management}
\end{figure}

The agent maintains context as a DAG of \emph{conversation nodes}, one per turn (Figure~\ref{fig:context_management}). Each node records the model response, tool calls, observations, and per-call token usage from one step. Every node carries two sets of pointers: a mutable \texttt{parent\_id} that defines the current \emph{active chain} sent to the LLM, and an immutable \texttt{history\_prev}/\texttt{history\_next} pair that defines the \emph{full history} of original turns. The active chain is always a sub-graph of the full history.

As tasks grow longer, the accumulated short-term context can exceed the model's token budget. Existing remedies span token-level prompt compression~\cite{longllmlingua}, attention-sink-based KV retention for streaming inference~\cite{streamingllm}, and OS-style virtual context management for general LLM agents~\cite{memgpt}; positional studies further document significant degradation when relevant content is buried in the middle of a long context~\cite{lostinmiddle}, which motivates the explicit first/last pinning we adopt below. To handle this at the agent level, MUSE applies adaptive context compression with two levels. \textbf{Level-1} (single-node compression) scans the active chain for individual nodes whose token footprint exceeds a per-node threshold (typically a large tool output or a verbose observation) and replaces that node's content with a compact summary while keeping it in the chain. If the total context is still over budget after Level-1, \textbf{Level-2} (chain compression) merges a contiguous range of intermediate nodes into a single synthetic summary node, which then takes the place of those nodes in the active chain. We always try Level-1 first because it is the strictly less destructive operation: only the offending node's payload is rewritten, while the per-turn boundaries and the full plan/action/observation structure of the chain are preserved, so downstream turns can still reference earlier turns by their original positions. Level-2 collapses several turns into one synthetic node and loses that per-turn structure, so we invoke it only when single-node summaries alone cannot bring the chain under budget. In both levels the original nodes remain in the full history, so the active chain is always recoverable. Long-term memory and the skill bank, by contrast, are stored separately and are not subject to compression, allowing the agent to accumulate experience across sessions without loss.

In addition, the agent's full state, including conversation history, skill usage records, and execution metadata, is persisted as a snapshot after each session. This allows tasks to be resumed from an intermediate state without restarting from scratch, which is essential for complex, long-horizon workflows that may span multiple sessions.

\FloatBarrier
\section{Experiments}
\label{sec:experiments}
We first conduct experiments on SkillsBench to evaluate three aspects of our framework: whether skill usage improves agent performance, whether MUSE-Autoskill can automatically generate effective skills from its own experience, and whether generated skills can transfer across agents. We then evaluate the same agent family on SkillLearnBench to test whether the skill-use and skill-creation trends hold on an independent continual-skill benchmark.

\subsection{Experimental Setup}

\paragraph{Benchmarks}
We evaluate on two benchmarks designed for skill use and skill creation. \textbf{SkillsBench}~\cite{skillsbench} assesses AI agents on real-world tasks that require domain-specific knowledge and tool use. Each task runs in an isolated Docker container and is graded by an automated verifier that checks only the final output files, assigning a reward in $[0, 1]$. We use the 75-task common set listed in Appendix~\ref{app:tasks}, grouped into four super-domains: science \& engineering, data analysis, document processing, and ops \& planning.

We further evaluate on \textbf{SkillLearnBench}~\cite{skilllearnbench}, an independent benchmark designed for continual skill generation on real-world agent tasks. It contains 20 skill-dependent tasks across software engineering, information retrieval, productivity tools, data and analytics, content and creative work, and utility workflows. Each task has multiple verified instances, yielding 100 instances in total; each instance is also executed in an isolated Docker environment and graded by an automated verifier over the submitted artifacts.

\paragraph{Agents and Models}
We evaluate four agents, all using GPT-5.5 as the backbone model: \textbf{MUSE-Autoskill} (our method), \textbf{Codex}, \textbf{Claude Code}, and \textbf{Hermes}. Since all agents share the same underlying model, performance differences reflect agent system design (including tool strategies, context management, planning, and skill usage) rather than model capacity.

\paragraph{Model Backend}
In all experiments, Hermes, Codex, Claude Code, and MUSE-Autoskill use the same GPT-5.5 deployment, \texttt{gpt-5.5-2026-04-24} (reported as GPT-5.5 (04/24/2026)); in particular, Claude Code's model calls are routed to this GPT-5.5 deployment through a compatibility bridge. We did not override decoding parameters such as temperature or top-$p$, so provider defaults are used throughout. Thus, performance differences reflect the surrounding agent systems, including prompts, tool loops, context handling, and skill-loading mechanisms, rather than different model backbones.

\paragraph{Evaluation Protocol}
For SkillsBench, each agent--task--configuration combination is run 5 times independently in separate Docker containers. We report all-task accuracy over 375 runs: the sum of rewards divided by $75 \times 5$. For SkillLearnBench, each of the 100 verified instances is evaluated once, and accuracy is the number of successful instances divided by 100. In both benchmarks, self-created-skill settings use the same denominator as their corresponding no-skill and human-skill settings; tasks or instances without a generated skill contribute 0 rather than being backfilled with the no-skill result.

\subsection{Effect of Skill Usage}

\paragraph{Setup}
We compare each agent under two shared conditions: \textit{without skills} (the agent relies solely on its own knowledge) and \textit{with human skills} (benchmark-provided, human-authored skills are injected into the agent's workspace at task start). For Codex, Claude Code, and MUSE-Autoskill, Table~\ref{tab:skill_effect} also reports the headline self-created-skill result under the same benchmark denominator. Hermes is included only for the without-skill and human-skill comparison in SkillLearnBench.

\paragraph{Results}
Table~\ref{tab:skill_effect} summarizes the main results on both benchmarks. Human skills improve every agent on both benchmarks. MUSE-Autoskill achieves the highest no-skill, human-skill, and self-created-skill accuracy in both settings, reaching 59.67\% with human skills on SkillsBench and 72.0\% on SkillLearnBench.

\begin{table}[t]
\caption{Main accuracy results (\%) on SkillsBench and SkillLearnBench. SkillsBench uses the strict 75-task $\times$ 5-run denominator; SkillLearnBench uses 100 verified instances. Self-created skills are evaluated only for self-creating agents. \textbf{Bold} = best within each benchmark column; \colorbox{autoskillblue}{blue rows} = MUSE-Autoskill (ours).}
\centering
\footnotesize
\setlength{\tabcolsep}{4pt}
\fitautotable{%
\begin{tabular}{llccc}
\toprule
\textbf{Benchmark} & \textbf{Agent} & \textbf{Without Skills} & \textbf{Human Skills} & \textbf{Self-Created Skills} \\
\midrule
SkillsBench & Hermes & 37.24\% & 48.02\% & -- \\
SkillsBench & Codex & 44.80\% & 57.58\% & 47.52\% \\
SkillsBench & Claude Code & 42.43\% & 56.15\% & 44.27\% \\
\rowcolor{autoskillblue} SkillsBench & MUSE-Autoskill (Ours) & \textbf{46.95\%} & \textbf{59.67\%} & \textbf{53.42\%} \\
\midrule
SkillLearnBench & Hermes & 37.0\% & 70.0\% & -- \\
SkillLearnBench & Codex & 39.0\% & 68.0\% & 40.0\% \\
SkillLearnBench & Claude Code & 33.0\% & 63.0\% & 37.0\% \\
\rowcolor{autoskillblue} SkillLearnBench & MUSE-Autoskill (Ours) & \textbf{43.0\%} & \textbf{72.0\%} & \textbf{48.0\%} \\
\bottomrule
\end{tabular}
}

\label{tab:skill_effect}
\end{table}

\paragraph{Discussion}
The consistent improvement across agents suggests that the skill mechanism itself is effective under strict denominators that count operational failures as zero. MUSE-Autoskill leads both no-skill and human-skill settings, while Codex and Claude Code are close when human skills are available. SkillLearnBench shows a larger human-skill lift than SkillsBench because it is explicitly built around skill-dependent instances, but it also shows the same automatic skill-generation bottleneck: self-created skills improve over the no-skill condition, yet remain below human-authored skills.

\paragraph{Per-Domain Breakdown}
SkillsBench's per-task \texttt{category} metadata is free-text, so we group the 75 common-set tasks into four super-domains: \textbf{Science \& Engineering} (18 tasks), \textbf{Data Analysis} (18), \textbf{Document Processing} (14), and \textbf{Ops \& Planning} (25). Appendix Table~\ref{tab:task_list} lists the task-level category and assigned super-domain for every task. Figure~\ref{fig:per_domain} and Table~\ref{tab:per_domain} report accuracy in each domain under both conditions. MUSE-Autoskill achieves the highest human-skill score in 3 of 4 domains and overall; Claude Code is strongest in Data Analysis.

\begin{table}[t]
\caption{Per-domain accuracy (\%) under \textit{without skills} (w/o) and \textit{with human skills} (w/ hum) conditions on the 75-task common set. \textbf{Bold} = best in each row's human-skill columns; \colorbox{autoskillblue}{blue columns} = MUSE-Autoskill (ours).}
\centering
\small
\setlength{\tabcolsep}{3pt}
\fitautotable{%
\begin{tabular}{lcccccccccc}
\toprule
& & \multicolumn{2}{c}{\textbf{Hermes}} & \multicolumn{2}{c}{\textbf{Codex}} & \multicolumn{2}{c}{\textbf{Claude Code}} & \multicolumn{2}{c}{\cellcolor{autoskillblue}\textbf{MUSE-Autoskill (Ours)}} & \\
\cmidrule(lr){3-4}\cmidrule(lr){5-6}\cmidrule(lr){7-8}\cmidrule(lr){9-10}
\textbf{Domain} & \textbf{\#} & w/o & w/ hum & w/o & w/ hum & w/o & w/ hum & \cellcolor{autoskillblue}w/o & \cellcolor{autoskillblue}w/ hum & \textbf{Best} \\
\midrule
Science \& Engineering & 18 & 40.61 & 58.89 & 54.52 & 67.12 & 52.33 & 63.88 & \cellcolor{autoskillblue}54.57 & \cellcolor{autoskillblue}\textbf{67.97} & \textbf{Ours} \\
Data Analysis          & 18 & 36.34 & 40.60 & 38.31 & 50.18 & 38.05 & \textbf{52.59} & \cellcolor{autoskillblue}39.49 & \cellcolor{autoskillblue}51.48 & Claude Code \\
Document Processing    & 14 & 54.29 & 58.57 & 68.57 & 72.86 & 62.86 & 64.29 & \cellcolor{autoskillblue}67.14 & \cellcolor{autoskillblue}\textbf{74.29} & \textbf{Ours} \\
Ops \& Planning        & 25 & 25.92 & 39.64 & 29.16 & 47.48 & 27.00 & 48.60 & \cellcolor{autoskillblue}35.52 & \cellcolor{autoskillblue}\textbf{51.40} & \textbf{Ours} \\
\midrule
\textbf{Overall} & 75 & 37.24 & 48.02 & 44.80 & 57.58 & 42.43 & 56.15 & \cellcolor{autoskillblue}46.95 & \cellcolor{autoskillblue}\textbf{59.67} & \textbf{Ours} \\
\bottomrule
\end{tabular}
}

\label{tab:per_domain}
\end{table}

\subsection{Automatic Skill Generation}
\label{sec:skill_gen}

\paragraph{Setup}
We analyze automatic skill generation in detail on the 75-task SkillsBench common set. The process follows a two-phase protocol. In \textbf{Phase~1}, the agent solves each task without skills. For tasks where a usable source trajectory is available, we invoke the agent's skill-creation mechanism to distill it into a \texttt{SKILL.md} and optional helper scripts. In \textbf{Phase~2}, the generated skill is injected back and the same agent is re-evaluated for 5 runs. Tasks without a usable generated skill are not dropped; they are counted as 0 in the 75-task denominator.

\paragraph{Results}
Table~\ref{tab:skill_gen} compares self-created skill performance for Codex, Claude Code, and MUSE-Autoskill. Self-created skills improve all three agents under the strict all-75 metric. MUSE-Autoskill obtains the largest gain and the strongest self-created result, improving from 46.95\% to \textbf{53.42\%}.

\begin{table}[t]
\caption{Self-created skill performance on the 75-task common set. Uncovered tasks are counted as 0 in all-task accuracy. Covered-task accuracy is reported only to diagnose skill quality conditional on successful generation. \textbf{Bold} = best in column; \colorbox{autoskillblue}{blue row} = MUSE-Autoskill (ours).}
\centering
\footnotesize
\setlength{\tabcolsep}{3.5pt}
\fitautotable{%
\begin{tabular}{lrrrrr}
\toprule
\textbf{Agent} & \textbf{Covered} & \textbf{Uncovered} & \textbf{All-75 Acc.} & \textbf{Covered Acc.} & \textbf{Lift vs.\ w/o} \\
\midrule
Codex & \textbf{47} & 28 & 47.52\% & 75.83\% & +2.72 pp \\
Claude Code & 44 & 31 & 44.27\% & 75.45\% & +1.84 pp \\
\rowcolor{autoskillblue} MUSE-Autoskill (Ours) & \textbf{47} & 28 & \textbf{53.42\%} & \textbf{85.24\%} & \textbf{+6.47 pp} \\
\bottomrule
\end{tabular}
}

\label{tab:skill_gen}
\end{table}

\paragraph{Discussion}
On covered tasks, generated skills are highly effective: MUSE-Autoskill reaches 85.24\%, above its 81.17\% human-skill accuracy on the same subset, while Codex and Claude Code reach 75.83\% and 75.45\%. This suggests that once a successful trajectory can be converted into a reusable skill, the resulting lifecycle-managed skill can match or exceed benchmark-provided human guidance. The all-task scores are lower because 28--31 tasks still have no usable generated skill and therefore contribute 0. The primary bottleneck is therefore \emph{coverage}: generating usable skills for more tasks, not only improving the skills that already exist.

\subsection{Cross-Agent Skill Transfer}

\paragraph{Setup}
We test whether generated skills can benefit a different agent. We inject skills created by Codex, Claude Code, and MUSE-Autoskill into Hermes without task-specific modification, and evaluate Hermes on the 75-task common set. The source generated-skill banks cover 47 Codex-created-skill tasks, 44 Claude Code-created-skill tasks, and 47 MUSE-Autoskill-created-skill tasks. Tasks outside the corresponding source-skill coverage are counted as 0, matching the strict all-task scoring protocol above.

\paragraph{Results}
Table~\ref{tab:transfer} summarizes the results. Here, ``Covered'' is the number of source generated-skill tasks used for transfer, while ``Uncovered'' is the number of remaining tasks counted as 0. MUSE-Autoskill-created skills transfer best to Hermes: they raise Hermes from 37.24\% without skills to \textbf{51.90\%}, which is also above Hermes with human skills at 48.02\%.

\begin{table}[t]
\caption{Cross-agent transfer into Hermes on the 75-task common set. Covered tasks use transferred-skill runs; uncovered, missing, and error runs are counted as 0. \textbf{Bold} = best non-baseline transfer result.}
\centering
\footnotesize
\setlength{\tabcolsep}{3.5pt}
\fitautotable{%
\begin{tabular}{lrrrr}
\toprule
\textbf{Hermes Configuration} & \textbf{Covered} & \textbf{Uncovered} & \textbf{All-75 Acc.} & \textbf{Delta vs.\ w/o} \\
\midrule
Without skills & 75 & 0 & 37.24\% & -- \\
Human skills & 75 & 0 & 48.02\% & +10.78 pp \\
With Codex-created skills & 47 & 28 & 37.01\% & -0.23 pp \\
With Claude Code-created skills & 44 & 31 & 45.97\% & +8.73 pp \\
\rowcolor{autoskillblue} With MUSE-Autoskill-created skills & 47 & 28 & \textbf{51.90\%} & \textbf{+14.66 pp} \\
\bottomrule
\end{tabular}
}

\label{tab:transfer}
\end{table}

\paragraph{Discussion}
MUSE-Autoskill-created skills transfer most effectively under the all-75 metric, improving Hermes by +14.66 pp and exceeding the human-skill condition by 3.88 pp. Claude Code-created skills also improve Hermes to 45.97\%, while Codex-created skills reach 37.01\% and do not improve over the no-skill baseline in this run. Transfer is therefore a joint function of skill representation quality and coverage; MUSE-Autoskill and Codex have the same 47-task source coverage here, but MUSE-Autoskill-created skills are substantially more effective when transferred.

\paragraph{Skill Generation and Usage Cost}
Table~\ref{tab:skill_cost} compares the one-time cost of generating a skill and the per-task cost of reusing it for Codex, Claude Code, and MUSE-Autoskill. Rows are computed on each agent's own covered subset (47 Codex tasks, 44 Claude Code tasks, and 47 MUSE-Autoskill tasks), so the table diagnoses reuse economics rather than replacing the all-75 accuracy metric in Table~\ref{tab:skill_gen}. MUSE-Autoskill has the lowest one-time token cost (364K) and is the only agent whose self-created skills exceed its human-skill covered-subset accuracy while reducing both median tokens and latency. Claude Code self-created skills are cheaper but slightly less accurate than human skills, while Codex self-created skills are faster but substantially more token-heavy. The final block is a transfer diagnostic: Hermes remains the Hermes agent, but uses the MUSE-Autoskill-created skill bank on the MUSE-covered subset; token totals are omitted for these Hermes transfer rows because the current logs do not expose normalized token totals.

\begin{table}[t]
\caption{Self-created skill generation and usage cost. Self-created-agent rows use each source agent's own covered subset; Hermes rows use the 47-task MUSE-covered subset. Tokens are median per-run totals when parseable; Hermes transfer-row token totals are omitted because the current logs do not expose normalized token totals. \colorbox{autoskillblue}{Blue rows} = MUSE-Autoskill-created skills (ours).}
\centering
\small
\setlength{\tabcolsep}{4pt}
\fitautotable{%
\begin{tabular}{llrrrrr}
\toprule
\textbf{Agent} & \textbf{Configuration} & \textbf{Covered} & \textbf{Acc.} & \textbf{Tokens} & \textbf{Latency (s)} & \textbf{Turns} \\
\midrule
\multicolumn{7}{l}{\emph{Self-created skill lifecycle cost (source agent's covered subset)}} \\
Codex & One-time skill creation & 47 & -- & 446K & 229.4 & 20 \\
Codex & Without skills & 47 & 71.5\% & 325K & 944.5 & 12 \\
Codex & Human skills & 47 & 76.6\% & 365K & 794.7 & 14 \\
Codex & Self-created skills & 47 & 75.8\% & 922K & 423.1 & 25 \\
\midrule
Claude Code & One-time skill creation & 44 & -- & 557K & 147.2 & 16 \\
Claude Code & Without skills & 44 & 71.3\% & 503K & 294.0 & 18 \\
Claude Code & Human skills & 44 & 77.8\% & 625K & 248.7 & 21 \\
Claude Code & Self-created skills & 44 & 75.5\% & \textbf{352K} & \textbf{159.0} & \textbf{14} \\
\midrule
\rowcolor{autoskillblue} MUSE-Autoskill & One-time skill creation & 47 & -- & \textbf{364K} & 156.3 & \textbf{6} \\
\rowcolor{autoskillblue} MUSE-Autoskill & Without skills & 47 & 74.9\% & 579K & 729.3 & 20 \\
\rowcolor{autoskillblue} MUSE-Autoskill & Human skills & 47 & 81.2\% & 638K & 755.7 & 20 \\
\rowcolor{autoskillblue} MUSE-Autoskill & Self-created skills (ours) & 47 & \textbf{85.2\%} & 499K & 434.7 & 15 \\
\midrule
\multicolumn{7}{l}{\emph{Transfer usage on the MUSE-covered subset}} \\
Hermes & Without skills & 47 & 59.0\% & -- & 336.0 & 14 \\
Hermes & Human skills & 47 & 64.3\% & -- & 342.9 & 13 \\
\rowcolor{autoskillblue} Hermes & With MUSE-created skills (ours) & 47 & \textbf{82.8\%} & -- & \textbf{262.9} & \textbf{13} \\
\bottomrule
\end{tabular}
}
\label{tab:skill_cost}
\end{table}

\subsection{Generated Skill Capabilities}

\begin{figure}[t]
\centering
\includegraphics[width=\textwidth]{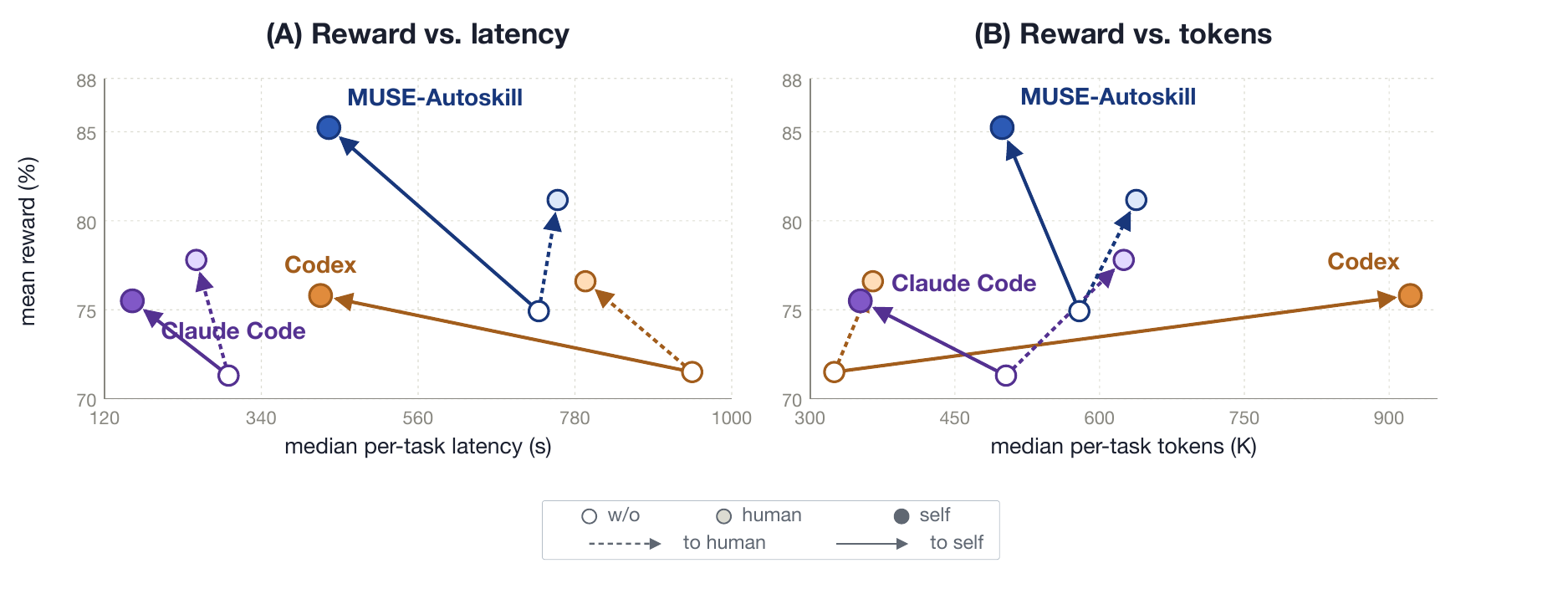}
\vspace{-.2in}
\caption{\textbf{Generated-skill tradeoffs on covered tasks.} Mean reward and median cost for Codex, Claude Code, and MUSE-Autoskill on each agent's own self-created-skill covered subset. \textbf{(A)} Reward vs.\ median per-task latency. \textbf{(B)} Reward vs.\ median per-task tokens. Open circle = without skills; light fill = human skills; solid fill = self-created skills. Dashed arrows point from no skills to human skills; solid arrows point from no skills to self-created skills.}
\label{fig:gen_skill_lift}
\vspace{-.1in}
\end{figure}

\paragraph{Aggregate Performance}
Figure~\ref{fig:gen_skill_lift} plots mean reward against cost for Codex, Claude Code, and MUSE-Autoskill on each agent's own self-created-skill covered subset. Each arrow starts from the no-skill condition: dashed arrows point to the human-skill condition, while solid arrows point to the self-created-skill condition. MUSE-Autoskill shows the strongest pattern: self-created skills improve covered-task reward from 74.92\% without skills and 81.17\% with human skills to 85.24\%, while reducing median latency to 434.7\,s and median tokens to 499K. Claude Code self-created skills are cheaper than its human-skill condition, but slightly lower in reward (75.45\% vs.\ 77.75\%). Codex self-created skills improve over its no-skill baseline and cut latency sharply, but they consume more tokens than both no-skill and human-skill runs. The all-75 score remains lower because generated-skill coverage is incomplete; this is the coverage bottleneck discussed below.

\paragraph{Case Studies}
We highlight three skills where the generated artifact carries non-trivial domain knowledge, plus one regression.

\noindent\textit{(i) \texttt{adaptive-cruise-control}} requires a discrete PID controller satisfying verifier constraints on overshoot, steady-state error, and rise time. MUSE-Autoskill without skills achieves 40\% (2 of 5 runs). The generated skill \texttt{implement-acc-simulation} codifies the discrete PID equation, anti-windup, gain-tuning heuristics, and the JSON file format required by the verifier; self-created accuracy reaches \textbf{100\%}. Hermes using the same MUSE-created skill improves from 20\% to 60\%, confirming that the skill transfers domain knowledge rather than memorizing the MUSE runtime.

\noindent\textit{(ii) \texttt{flink-query}} asks the agent to author an Apache Flink Java job that reads gzipped Google ClusterData traces, performs microsecond event-time sessionization, and emits tuples in an exact format. The baseline solves only one of five runs (20\%) because the agent cannot recover the project's POJO and \texttt{AppBase} skeleton conventions from documentation alone within its turn budget. The generated skill \texttt{implement-\allowbreak clusterdata-\allowbreak flink-\allowbreak session-\allowbreak query} packages the schema parsing, the \texttt{clusterdata.\allowbreak utils.\allowbreak AppBase} extension protocol, event-time session triggers, and a Maven-based validation recipe with synthetic gzipped data; Phase~2 reaches \textbf{100\%} across all five runs.

\noindent\textit{(iii) \texttt{weighted-gdp-calc}} requires filling an Excel workbook with two-condition lookups and \texttt{SUMPRODUCT}-based weighted means while preserving existing formatting and avoiding macros/VBA. The generated skill \texttt{excel-financial-formula-modeling} names \texttt{openpyxl} as the right tool, lists the formula patterns, and adds a verification step that recomputes target cells from source data; the baseline jumps from 20\% to \textbf{100\%}. Notably, the same skill description guides Hermes through the identical workflow without modification.

\noindent\textit{(iv) Regression: \texttt{hvac-control}.} The largest MUSE self-created regression (80\% $\to$ 20\%) occurs on a task that requires PI control of a first-order thermal simulator. The source trajectory used a calibration window and gain-estimation routine specific to that simulator's noise profile; when re-applied in fresh runs, the variance in calibration data occasionally produces tuned gains outside the verifier's stability margin. This is a case where the skill encodes a procedure that worked once but is \emph{less robust} than baseline trial-and-error, and motivates the audit finding (next subsection) that some skills carry source-trajectory-specific assumptions that limit out-of-distribution robustness.

\subsection{Analysis}

\paragraph{Skill Quality Audit}
We manually inspect the 50 MUSE-generated skill packages covering 47 tasks for potential benchmark leakage. None of the inspected skills hardcode expected verifier outputs, branch on task identifiers, or read from ground-truth files. A subset of skills contain benchmark-specific assumptions such as fixed file names, directory paths, or numerical ranges derived from the source run. These do not constitute cheating but may limit generalization to out-of-distribution inputs. The structural distribution of agent-created and human-authored skill packages is summarized in Table~\ref{tab:skill_anatomy_main}.

\begin{figure}[t]
\centering
\includegraphics[width=\textwidth]{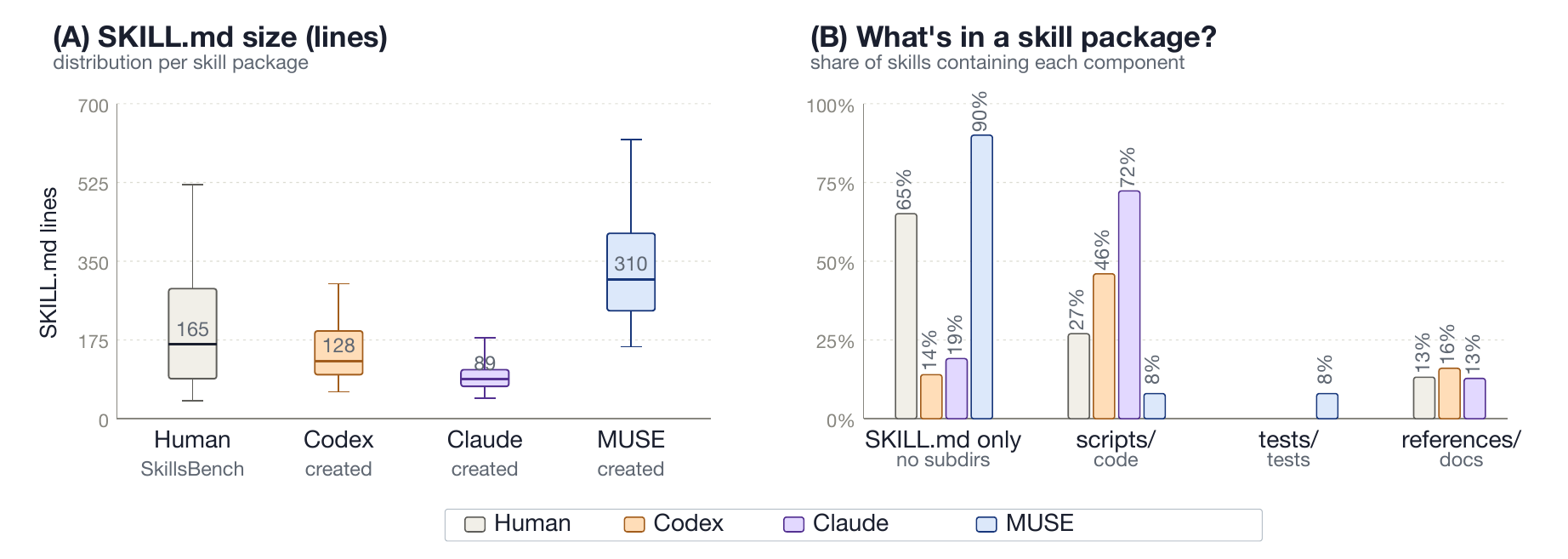}
\vspace{-.2in}
\caption{\textbf{Skill anatomy: human-authored and agent-created packages.} \textbf{(A)} \texttt{SKILL.md} line counts on the 75-task common set: MUSE-Autoskill-created skills are longest (median 310 lines), compared with human-authored SkillsBench skills (165), Codex-created skills (128), and Claude Code-created skills (89). \textbf{(B)} Share of skill packages containing each component. Codex and Claude Code more often emit helper \texttt{scripts/}; MUSE-Autoskill more often emits self-contained \texttt{SKILL.md}-only packages and is the only group with generated \texttt{tests/} packages in this audit.}
\label{fig:skill_anatomy}
\end{figure}

\paragraph{Skill Anatomy and Distribution}
Figure~\ref{fig:skill_anatomy} and Table~\ref{tab:skill_anatomy_main} compare agent-created skill packages from Codex, Claude Code, and MUSE-Autoskill against 189 human-authored SkillsBench packages from the same 75-task common set. MUSE-Autoskill-created skills are longer than the other generated baselines and human-authored skills, but the extra length is concentrated in procedural detail: input/output schemas, verifier-facing file conventions, failure modes, and validation steps. Codex and Claude Code more frequently bundle helper scripts, whereas MUSE-Autoskill more often distills a self-contained textual procedure and uniquely generates test packages in this audit.

\begin{table}[t]
\caption{Skill package anatomy on the 75-task common set. Line count and size are medians over \texttt{SKILL.md}; directory shares are package percentages and are not mutually exclusive.}
\label{tab:skill_anatomy_main}
\centering
\footnotesize
\setlength{\tabcolsep}{3.2pt}
\fitautotable{%
\begin{tabular}{lrrrrrrrrr}
\toprule
\textbf{Source} & \textbf{Packages} & \textbf{Tasks} & \textbf{Lines} & \textbf{IQR} & \textbf{Size} & \textbf{Only} & \textbf{scripts/} & \textbf{tests/} & \textbf{refs/} \\
 & & & \textbf{median} & & \textbf{KB} & \textbf{SKILL.md} & & & \\
\midrule
Human-authored SkillsBench & 189 & 75 & 165 & 89--289 & 5.2 & 65.1\% & 27.0\% & 0.0\% & 13.2\% \\
Codex-created & 50 & 47 & 128 & 98--195 & 7.0 & 14.0\% & 46.0\% & 0.0\% & 16.0\% \\
Claude Code-created & 47 & 44 & 89 & 72--109 & 5.6 & 19.1\% & 72.3\% & 0.0\% & 12.8\% \\
\rowcolor{autoskillblue} MUSE-Autoskill-created & 50 & 47 & \textbf{310} & 240--412 & \textbf{13.2} & \textbf{90.0\%} & 8.0\% & \textbf{8.0\%} & 0.0\% \\
\bottomrule
\end{tabular}
}
\end{table}

\begin{figure}[t]
\centering
\includegraphics[width=\textwidth]{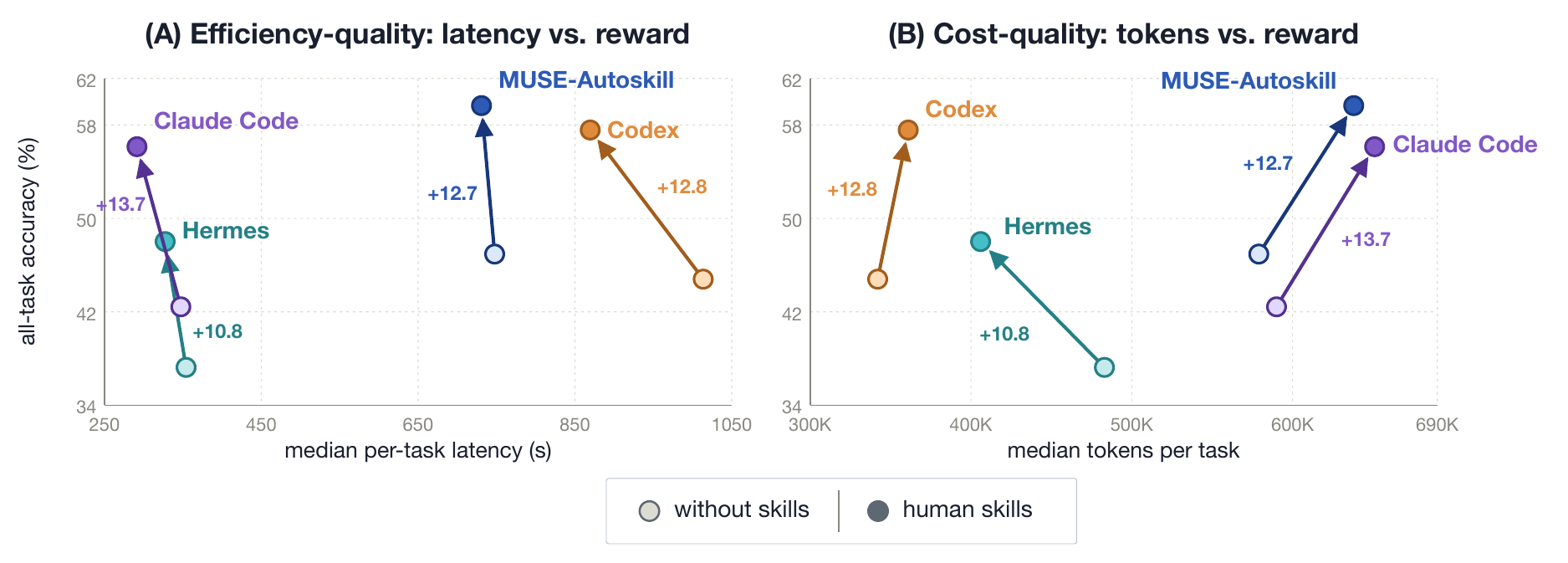}
\vspace{-.2in}
\caption{\textbf{Skill-induced tradeoffs in two dimensions} on the 75-task common set. \textbf{(A) Latency vs.\ reward.} Human skills move all four agents upward in reward and reduce median latency. \textbf{(B) Tokens vs.\ reward.} For agents with parseable total-token traces, human skills increase reward with modest token overhead for Codex, Claude Code, and MUSE-Autoskill, while Hermes total tokens decrease on its available summary traces. Colors match Figure~\ref{fig:per_domain}.}
\label{fig:tradeoff_panels}
\vspace{-.1in}
\end{figure}

\paragraph{Reward--Cost Relationship}
Human skills raise reward for every agent while reducing median wall-clock latency (Figure~\ref{fig:tradeoff_panels}A). This is the main cost-quality pattern: skills add context, but they also replace exploratory reasoning with a more direct procedure. Turn counts do not uniformly decrease, so the robust claim is latency improvement rather than fewer ReAct steps. Hermes, Codex, Claude Code, and MUSE-Autoskill all move to higher reward and lower median latency under human skills; full latency percentiles for all agents are reported in Appendix~\ref{app:latency}.

\paragraph{Token Usage}
Token accounting is diagnostic because normalized fresh/cache traces are not available for every runtime. For the agents with parseable totals, skills increase reward while changing the median token footprint in different directions (Figure~\ref{fig:tradeoff_panels}B and Table~\ref{tab:token_reward_main}). Codex gains +12.78 pp for a +5.4\% median-token increase. MUSE-Autoskill gains +12.72 pp for a +10.1\% increase, with most input tokens served from prompt cache in both conditions (Appendix~\ref{app:tokens}). Hermes exposes total-token summaries for a subset of runs and decreases from 483K to 406K median tokens under human skills; Claude Code exposes total-token metadata but not the same fresh/cache split.

\begin{table}[t]
\caption{Token usage diagnostic on the 75-task common set. Totals are median per-run tokens where parseable. Fresh/cache decomposition is reported separately in Appendix~\ref{app:tokens}.}
\label{tab:token_reward_main}
\centering
\small
\setlength{\tabcolsep}{4pt}
\fitautotable{%
\begin{tabular}{lrrr}
\toprule
\textbf{Agent} & \textbf{Total w/o} & \textbf{Total human} & \textbf{$\Delta$ total} \\
\midrule
Hermes & 483K & 406K & $-$16.0\% \\
Codex & 342K & 361K & +5.4\% \\
Claude Code & 590K & 651K & +10.3\% \\
\rowcolor{autoskillblue} MUSE-Autoskill & 579K & 638K & +10.1\% \\
\bottomrule
\end{tabular}
}
\end{table}

\begin{figure}[t]
\centering
\includegraphics[width=\textwidth]{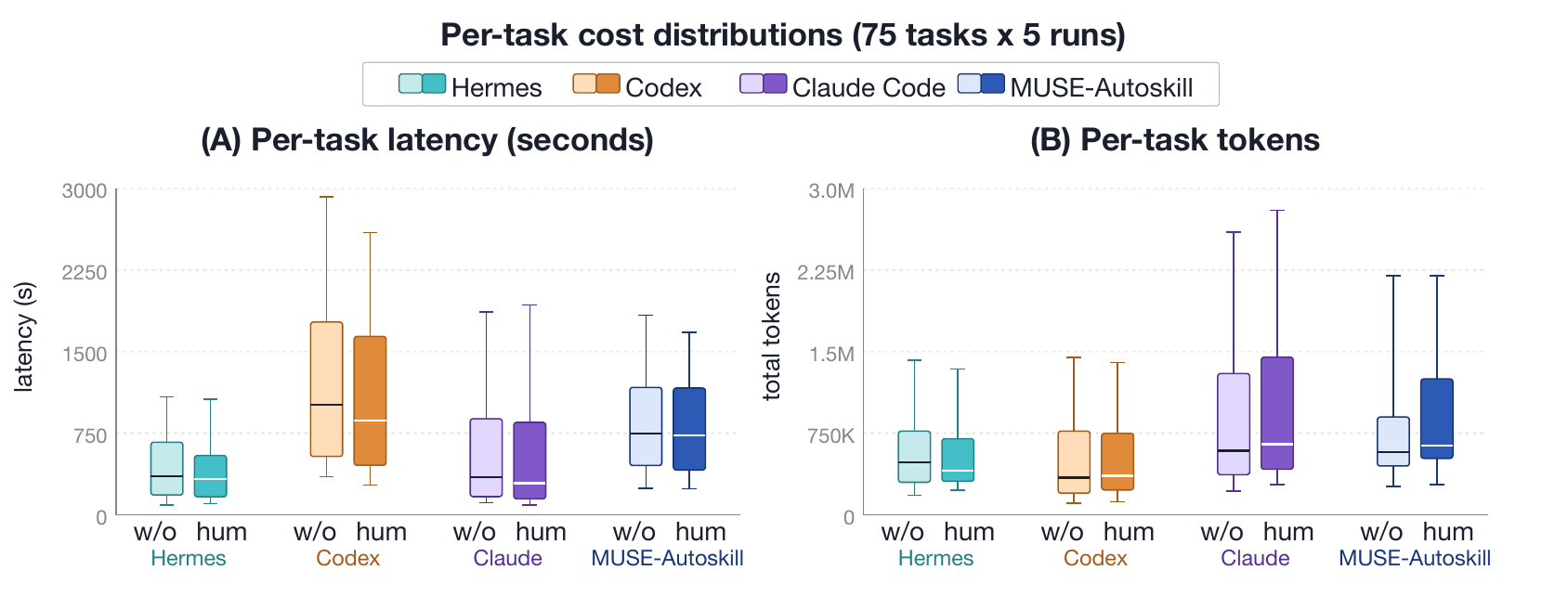}
\vspace{-.25in}
\caption{\textbf{Per-task cost distributions} on the 75-task common set. \textbf{(A)} Agent execution latency in seconds for Hermes, Codex, Claude Code, and MUSE-Autoskill. \textbf{(B)} Total tokens for Hermes, Codex, Claude Code, and MUSE-Autoskill, where parseable token traces are available. Boxes span the IQR, the center line marks the median, and whiskers extend to p10 and p90 for each condition.}
\label{fig:cost_distributions}
\end{figure}

\paragraph{Cost Distributions}
Figure~\ref{fig:cost_distributions} shows the full per-run latency and token distributions behind the medians in Table~\ref{tab:token_reward_main} and Appendix Table~\ref{tab:latency_dist}. Hermes and Claude Code have the lowest median latency, while Codex has the largest latency tail. MUSE-Autoskill uses more tokens than Codex and Claude Code because its loop carries the skill lifecycle and memory machinery, but the generated-skill subset in Table~\ref{tab:skill_cost} shows that a distilled skill can reduce this cost sharply once it exists.

\paragraph{Generated-Skill Cost Amortization}
On the 47 tasks where MUSE-Autoskill creates a usable skill, the generated skill is not just more accurate; it is also cheaper to use than both the no-skill and human-skill conditions (Table~\ref{tab:skill_cost}). Accuracy rises from 74.92\% without skills and 81.17\% with human skills to 85.24\% with self-created skills, while median latency drops to 434.7\,s and median token usage drops to 499K. Creating the skill is a one-time cost: median 363.6K tokens and 156.3\,s per covered task. Relative to human skills, the generated skill saves about 139K tokens and 321\,s per reuse, so the token cost breaks even after roughly three reuses and the latency cost after the first reuse.

\paragraph{Bottleneck Analysis}
MUSE-Autoskill leaves 28 of 75 tasks uncovered in self-creation. These are concentrated in Ops \& Planning (12 tasks) and Data Analysis (9), with smaller clusters in Science \& Engineering (5) and Document Processing (2). This pattern supports the central diagnosis from Table~\ref{tab:skill_gen}: generated skills are strong when a successful source trajectory exists, but Phase~1 exploration still fails to produce a reusable trajectory for many difficult tasks. Future work should therefore focus on improving source-trajectory coverage and on extracting partial diagnostic skills from failed trajectories rather than waiting for a fully successful run.

\section{Real-World Deployment and Impact}
\label{sec:impact}

Beyond the controlled SkillsBench evaluation, the skill-centric design of MUSE-Autoskill is already being adopted in production systems, where skills serve as the common unit of capability shared across agents and users. SkillMarket exposes the skill-creation pipeline to end users, distilling a successful trajectory into a reusable, self-tested skill package without manual authoring; planned releases add skill management and updating, so that deployed skills can be versioned and refined as tasks and environments drift over time. ArkClaw integrates the skill-retrieval component as a \emph{find-skill} capability, letting an agent locate the most relevant existing skill before synthesizing a new one, and a planned extension treats an entire agent as an invocable sub-agent, so that a single skill can encapsulate delegated multi-agent behavior.

SkillHub operationalizes the full skill lifecycle, covering creation, evaluation, memory, management, and refinement, as a hosted service that gives teams one place to store, evaluate, and govern skills together with their accumulated per-skill experience. Taken together, these deployments show that the lifecycle abstraction is not specific to our benchmark setting: the same retrieve-or-create decision, bundled tests, and per-skill memory carry over to systems built and used by different teams, and an improvement to a shared skill propagates to every agent and product that depends on it.

Looking forward, we expect skills to take on a broader role as the primitive for defining workflows. Rather than hand-wiring agent pipelines, developers will compose and version skills whose bundled tests and memory keep the resulting workflows self-documenting and easier to maintain. This shifts ongoing maintenance cost from bespoke glue code to a shared, continuously evaluated skill ecosystem, and we view these early deployments as evidence that a unified skill lifecycle is a practical foundation for agents whose capabilities compound, rather than erode, as they are maintained at scale.

\section{Conclusion}

We present a skill-centric agent framework that improves task-solving by acquiring, reusing, and refining skills through a unified lifecycle. By packaging reusable procedures as structured skills, the agent can avoid rediscovering the same commands, file formats, and validation steps on later runs. MUSE-Autoskill integrates skill creation, evaluation, execution, memory, and management around minimal built-in skills such as \textit{skill\_create} and \textit{web\_search}. On SkillsBench, it achieves the strongest human-skill accuracy (59.67\%), the strongest all-task self-created-skill result (53.42\%), and the best transfer into Hermes (51.90\%). On SkillLearnBench, it again leads the compared self-creating agents on 100 verified instances. Together with production deployments of skill creation, discovery, and lifecycle management, these results support skill packages as a practical unit for accumulating and reusing agent experience across repeated tasks.

\section*{Limitations}

Our evaluation covers the 75 SkillsBench tasks that all four compared runtimes can run locally, rather than the full 94-task benchmark; the excluded tasks often have more complex Docker environments and may be harder. Self-created skill coverage is incomplete: MUSE-Autoskill and Codex produce usable skills for 47 of 75 tasks, while Claude Code covers 44. Each covered-task skill is distilled from a single source trajectory, transfer is evaluated only into Hermes, and with 5 runs per task, individual-task confidence intervals remain wide. Full pairwise transfer among all four agents also remains untested.

A further concern is that each self-created skill is generated from one successful Phase~1 trajectory and re-evaluated on the \emph{same task}. Although the verifier is deterministic and no task-specific ground truth is fed into the skill (Section~\ref{sec:experiments}), this protocol may overstate within-task gains. SkillLearnBench uses a separate 100-instance, one-run-per-instance protocol and should be read as corroborating evidence rather than pooled with SkillsBench. Token traces are incomplete across runtimes, so token-cost results are diagnostic. Broader benchmarks are needed before treating the gains as domain-general.

A remaining risk is hallucination in generated skills or in the agent reasoning that invokes them. Skills can encode unsupported facts, brittle file-path or API assumptions, or heuristics overfit to a source trajectory. Unit tests, sandbox execution, verifier feedback, and leakage checks reduce but do not eliminate this risk, so high-impact deployments need stronger provenance tracking, adversarial tests, and human review before skills are shared across teams or exposed to user-facing workflows.

AI tools were used for drafting assistance, wording suggestions, and revision support. The research idea, system prototype, experimental design, experiment execution, result analysis, and final paper-level decisions were led by the human authors, who reviewed the AI-assisted text and take responsibility for the paper's content, claims, and conclusions.

\clearpage
\bibliographystyle{plainnat}
\bibliography{references}


\clearpage
\appendix

\section{Selected Task List}
\label{app:tasks}

Table~\ref{tab:task_list} lists the 75 SkillsBench tasks used for the four-agent comparison in Figure~\ref{fig:per_domain}. We keep the original SkillsBench category where available, falling back to the primary task tag when the category field is absent, and group tasks into the four super-domains used in our analysis: 18 Science \& Engineering, 18 Data Analysis, 14 Document Processing, and 25 Ops \& Planning tasks.

\begingroup
\small
\setlength{\tabcolsep}{3pt}
\begin{longtable}{rlll}
\caption{The 75-task SkillsBench common set used for the four-agent comparison.}
\label{tab:task_list}\\
\toprule
\textbf{\#} & \textbf{Task ID} & \textbf{SkillsBench Category} & \textbf{Super-domain} \\
\midrule
\endfirsthead
\toprule
\textbf{\#} & \textbf{Task ID} & \textbf{SkillsBench Category} & \textbf{Super-domain} \\
\midrule
\endhead
\midrule
\multicolumn{4}{r}{Continued on next page} \\
\endfoot
\bottomrule
\endlastfoot
 1 & \texttt{3d-scan-calc} & \texttt{engineering} & Science \& Engineering \\
 2 & \texttt{ada-bathroom-plan-repair} & \texttt{architecture} & Document Processing \\
 3 & \texttt{adaptive-cruise-control} & \texttt{control-systems} & Science \& Engineering \\
 4 & \texttt{azure-bgp-oscillation-route-leak} & \texttt{bgp-route} & Ops \& Planning \\
 5 & \texttt{bike-rebalance} & \texttt{transportation-logistics} & Ops \& Planning \\
 6 & \texttt{citation-check} & \texttt{research} & Document Processing \\
 7 & \texttt{civ6-adjacency-optimizer} & \texttt{games} & Ops \& Planning \\
 8 & \texttt{court-form-filling} & \texttt{document-processing} & Document Processing \\
 9 & \texttt{crystallographic-wyckoff-position-analysis} & \texttt{materials\_science} & Science \& Engineering \\
10 & \texttt{dapt-intrusion-detection} & \texttt{security} & Ops \& Planning \\
11 & \texttt{data-to-d3} & \texttt{Data Visualization} & Data Analysis \\
12 & \texttt{dialogue-parser} & \texttt{game} & Data Analysis \\
13 & \texttt{dynamic-object-aware-egomotion} & \texttt{video-analysis} & Science \& Engineering \\
14 & \texttt{earthquake-plate-calculation} & \texttt{geophysics} & Science \& Engineering \\
15 & \texttt{econ-detrending-correlation} & \texttt{economics} & Data Analysis \\
16 & \texttt{edit-pdf} & \texttt{pdf} & Document Processing \\
17 & \texttt{energy-market-pricing} & \texttt{energy} & Ops \& Planning \\
18 & \texttt{energy-unit-commitment} & \texttt{energy} & Ops \& Planning \\
19 & \texttt{enterprise-information-search} & \texttt{enterprise-search} & Ops \& Planning \\
20 & \texttt{exam-block-sequencing} & \texttt{scheduling} & Ops \& Planning \\
21 & \texttt{exceltable-in-ppt} & \texttt{Office Operation} & Document Processing \\
22 & \texttt{exoplanet-detection-period} & \texttt{astronomy} & Science \& Engineering \\
23 & \texttt{financial-modeling-qa} & \texttt{financial modeling} & Data Analysis \\
24 & \texttt{find-topk-similiar-chemicals} & \texttt{chemistry} & Science \& Engineering \\
25 & \texttt{fix-druid-loophole-cve} & \texttt{Security} & Ops \& Planning \\
26 & \texttt{fix-erlang-ssh-cve} & \texttt{erlang bugfix} & Ops \& Planning \\
27 & \texttt{flink-query} & \texttt{flink} & Ops \& Planning \\
28 & \texttt{flood-risk-analysis} & \texttt{data-processing} & Science \& Engineering \\
29 & \texttt{gravitational-wave-detection} & \texttt{astronomy} & Science \& Engineering \\
30 & \texttt{grid-dispatch-operator} & \texttt{energy} & Ops \& Planning \\
31 & \texttt{hvac-control} & \texttt{control-systems} & Science \& Engineering \\
32 & \texttt{invoice-fraud-detection} & \texttt{data-validation} & Data Analysis \\
33 & \texttt{jax-computing-basics} & \texttt{research} & Science \& Engineering \\
34 & \texttt{jpg-ocr-stat} & \texttt{data statistics} & Document Processing \\
35 & \texttt{lab-unit-harmonization} & \texttt{healthcare} & Data Analysis \\
36 & \texttt{lake-warming-attribution} & \texttt{data-processing} & Science \& Engineering \\
37 & \texttt{latex-formula-extraction} & \texttt{latex-extraction} & Document Processing \\
38 & \texttt{lean4-proof} & \texttt{formal method} & Science \& Engineering \\
39 & \texttt{llm-prefix-cache-replay} & \texttt{ml-systems} & Ops \& Planning \\
40 & \texttt{manufacturing-codebook-normalization} & \texttt{manufacturing} & Ops \& Planning \\
41 & \texttt{manufacturing-equipment-maintenance} & \texttt{manufacturing} & Ops \& Planning \\
42 & \texttt{manufacturing-fjsp-optimization} & \texttt{manufacturing} & Ops \& Planning \\
43 & \texttt{mars-clouds-clustering} & \texttt{data-science} & Science \& Engineering \\
44 & \texttt{multilingual-video-dubbing} & \texttt{multimodal-video-dubbing} & Document Processing \\
45 & \texttt{offer-letter-generator} & \texttt{document-generation} & Document Processing \\
46 & \texttt{organize-messy-files} & \texttt{file-management} & Ops \& Planning \\
47 & \texttt{paper-anonymizer} & \texttt{document-editing} & Document Processing \\
48 & \texttt{parallel-tfidf-search} & \texttt{Parallelization} & Data Analysis \\
49 & \texttt{pddl-tpp-planning} & \texttt{research} & Ops \& Planning \\
50 & \texttt{pdf-excel-diff} & \texttt{data-comparison} & Document Processing \\
51 & \texttt{pedestrian-traffic-counting} & \texttt{pedestrian traffic counting} & Science \& Engineering \\
52 & \texttt{powerlifting-coef-calc} & \texttt{data-analysis} & Data Analysis \\
53 & \texttt{pptx-reference-formatting} & \texttt{office-suite} & Document Processing \\
54 & \texttt{protein-expression-analysis} & \texttt{data-analysis} & Science \& Engineering \\
55 & \texttt{r2r-mpc-control} & \texttt{control-systems} & Science \& Engineering \\
56 & \texttt{radar-vital-signs} & \texttt{signal-processing} & Science \& Engineering \\
57 & \texttt{react-performance-debugging} & \texttt{web-performance} & Ops \& Planning \\
58 & \texttt{reserves-at-risk-calc} & \texttt{financial-analysis} & Data Analysis \\
59 & \texttt{sales-pivot-analysis} & \texttt{data-analysis} & Data Analysis \\
60 & \texttt{sec-financial-report} & \texttt{finance} & Data Analysis \\
61 & \texttt{shock-analysis-demand} & \texttt{financial-analysis} & Data Analysis \\
62 & \texttt{shock-analysis-supply} & \texttt{financial-analysis} & Data Analysis \\
63 & \texttt{simpo-code-reproduction} & \texttt{code reproduction} & Ops \& Planning \\
64 & \texttt{software-dependency-audit} & \texttt{security} & Ops \& Planning \\
65 & \texttt{spring-boot-jakarta-migration} & \texttt{Legacy Systems} & Ops \& Planning \\
66 & \texttt{syzkaller-ppdev-syzlang} & \texttt{security} & Ops \& Planning \\
67 & \texttt{taxonomy-tree-merge} & \texttt{ML/NLP} & Data Analysis \\
68 & \texttt{threejs-structure-parser} & \texttt{3d-graphics} & Data Analysis \\
69 & \texttt{threejs-to-obj} & \texttt{3d-graphics} & Data Analysis \\
70 & \texttt{tictoc-unnecessary-abort-detection} & \texttt{database-systems} & Ops \& Planning \\
71 & \texttt{travel-planning} & \texttt{travel-planning} & Ops \& Planning \\
72 & \texttt{video-silence-remover} & \texttt{media-processing} & Document Processing \\
73 & \texttt{video-tutorial-indexer} & \texttt{multimodal-processing} & Document Processing \\
74 & \texttt{weighted-gdp-calc} & \texttt{financial-analysis} & Data Analysis \\
75 & \texttt{xlsx-recover-data} & \texttt{spreadsheet} & Data Analysis \\
\end{longtable}
\endgroup

\section{FDABench Supplementary Evaluation}
\label{app:fdabench}

FDABench-Full~\cite{wang2025fdabench} evaluates agents on heterogeneous data-analysis tasks. It contains three task families: \emph{single-choice} questions, where the agent selects one option; \emph{multiple-choice} questions, where the agent returns a set of options under strict exact-match grading; and \emph{report} tasks, where the agent produces an analytical report from database-backed evidence. The full public benchmark contains 579 single-choice, 760 multiple-choice, and 668 report task instances. The underlying data instances span Spider2-lite, BIRD, DABStep, and Spider1-style databases, with easy, medium, and hard difficulty labels.

We include FDABench as a supplementary stress test rather than as a headline comparison. The MUSE-Autoskill score values correspond to the public FDABench data-agent leaderboard row listed as \emph{Data Analysis Agent (ByteDance Lark Base \& Hydra \& NovaBase Team)}. For choice tasks, FDABench reports execution accuracy (EX); MUSE-Autoskill obtains 74.10\% EX on single-choice tasks with 281.4s average latency and 179.6M aggregate tokens (310.2K per task), and 49.70\% EX on multiple-choice tasks with 183.2s average latency and 166.7M aggregate tokens (219.4K per task). For report tasks, FDABench reports rubric score (RS); MUSE-Autoskill obtains 85.20\% RS with 96.6s average latency.

\begin{figure}[H]
\centering
\includegraphics[width=\textwidth]{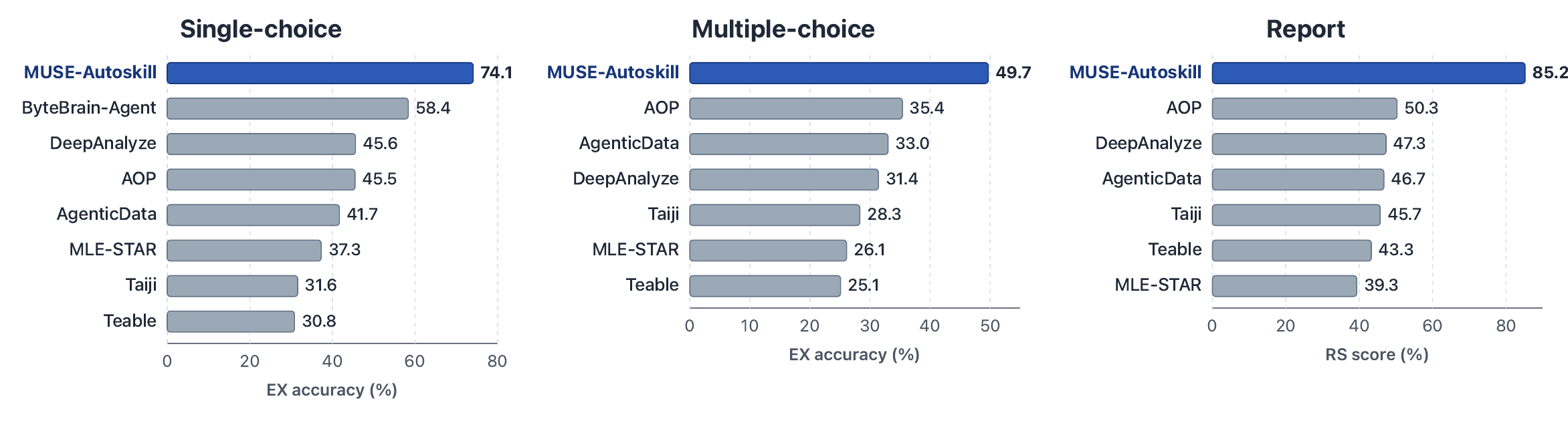}
\caption{FDABench public data-agent leaderboard comparison for single-choice, multiple-choice, and report tasks. The highlighted bars show MUSE-Autoskill; the score values correspond to the FDABench public row listed as \emph{Data Analysis Agent (ByteDance Lark Base \& Hydra \& NovaBase Team)}. Values are from the FDABench website's public \texttt{method\_aggregated.csv} file at \url{https://fdabench.github.io/static/data/method_aggregated.csv}.}
\label{fig:fdabench_leaderboard}
\end{figure}

\section{Skill Package Schema}
\label{app:schema}

A skill is a self-contained directory rooted at the kebab-case skill name. The directory always contains a top-level \texttt{SKILL.md} written in Markdown with a YAML frontmatter block, and may optionally contain the subdirectories \texttt{scripts/}, \texttt{tests/}, \texttt{resources/}, and \texttt{references/}. Skills that do not need code consist of \texttt{SKILL.md} alone, which is the dominant pattern in practice. The skill identifier is taken from the directory name, and the same name must appear in the frontmatter \texttt{name} field; this redundancy lets a skill be moved or copied without breaking its identity. The schema deliberately mirrors Anthropic's Agent Skills format~\cite{anthropic-skills} so that skills produced by MUSE-Autoskill can be loaded by any agent that already understands that format, without translation. The minimum viable skill file is:

\begin{verbatim}
---
name:        <kebab-case skill identifier; must match the directory name>
description: <one-paragraph natural-language description; this is what the
              agent reads when deciding whether to invoke the skill>
---

# <Skill title in Title Case>

## When to use
- Bullet list of triggering task types.

## Core principles
1. Numbered list of invariants the implementation must preserve.

## Recommended tools and libraries
- Concrete library names, CLI commands, or sandbox tools.

## Workflow
Step-by-step procedure the agent should follow at runtime.
\end{verbatim}

\paragraph{Catalog routing.} The frontmatter \texttt{description} field is the only piece of the skill that is surfaced eagerly: at the start of every task the runtime injects a YAML catalog of all available skills (each entry containing just \texttt{name} and \texttt{description}) into the agent's system prompt. The body of \texttt{SKILL.md} is loaded only after the agent decides, via the \texttt{read\_skill} tool, that the skill is worth pulling into context. This two-stage lookup keeps the per-call input cost flat in the size of the skill bank: a bank with 100 skills adds only $\sim$5--10K tokens of catalog, not the $\sim$500K tokens that loading every skill body would require.

\paragraph{Subdirectory conventions.} The optional subdirectories follow strict per-name conventions, so the agent can rely on layout when reading the skill at runtime: \texttt{scripts/} holds executable code (Python, shell, Node) that the skill instructs the agent to run inside the sandbox; \texttt{tests/} is used for pytest-compatible validation of code-backed skills and is absent for most text-only procedural skills; \texttt{resources/} holds passive auxiliary files (data tables, prompt fragments, reference documents) that the skill loads on demand at execution time; and \texttt{references/} (used by some human SkillsBench skills) holds reference documentation that the agent may read but is not expected to execute. When \texttt{tests/} is present, failed tests block registration; otherwise registration relies on the available sandbox/runtime checks described above. A skill never embeds dependencies: it relies on the sandbox image (or runtime \texttt{pip install} via the terminal tool) for any packages it needs, which keeps the skill bank itself a pure-text artifact safe to version-control and ship as a tarball.

\paragraph{Skill-level memory.} Alongside the on-disk skill, each skill gets a sibling \texttt{.memory.md} file (created lazily on first write) into which the agent appends notes, lessons, and usage observations across tasks. This file is the concrete realisation of the \emph{skill-level memory} described in Section~\ref{sec:skill_lifecycle}. It is intentionally outside the skill directory's published surface area (the leading dot, and exclusion from any \texttt{.tar} the user might ship) so that transferring the skill does \emph{not} transfer experience accumulated from prior runs; experience is per-agent.

\section{File-System Layout}
\label{app:filesystem}

This appendix documents the on-disk layout the agent assumes at runtime. Every path is configurable; we list the defaults so that an outside reader can understand where each component of a published trajectory came from in the run archive.

\paragraph{Agent home directory.} On the host the agent runtime defaults to \texttt{\$HOME/.autoskill} (overridable via the \texttt{AUTOSKILL\_HOME} environment variable). It is created on first launch and contains all persistent state that is not part of an individual session:

\begin{verbatim}
~/.autoskill/
+-- skills/                       # the skill bank: one directory per skill
|   +-- pdf-form-update-redaction/
|   |   +-- SKILL.md              # frontmatter + body
|   |   +-- .memory.md            # per-skill memory (Section 3.4)
|   |   +-- scripts/              # optional: executable code
|   |   +-- tests/                # optional: pytest-compatible
|   |   +-- resources/            # optional: data / docs
|   +-- csv-summarize/
|   |   +-- SKILL.md              # the typical "doc-only" skill
|   +-- ...
+-- memory/
|   +-- long_term_memory/
|       +-- memory.md             # cross-session notes, lessons learned
+-- sessions/                     # per-session workspaces (see below)
    +-- 2d9b1c67f73947c4863b26a45c5098a8/
    +-- ...
\end{verbatim}

\paragraph{Per-session workspace.} A new directory under \texttt{\$AUTOSKILL\_HOME/sessions/<session\_id>/} is created for each task invocation. The session id is a UUID-like string that also names the directory; the runtime persists the agent's complete state into this directory at the end of every session. Inside one session:

\begin{verbatim}
sessions/<session_id>/
+-- instruction.md                # the task prompt the agent received
+-- submitted_inputs/             # files supplied by the caller
+-- submitted_skillhub/           # any injected skills supplied at task start
+-- result_output_files/          # final artifacts the agent produced
+-- agent_message.md              # final-answer text returned to the caller
+-- agent.stdout.txt              # log stream incl. per-call token usage
+-- events.jsonl                  # one JSON event per tool call / turn
+-- memory.md                     # short-term (session-scoped) memory
+-- ctx_state.json                # serialised AgentContext (for resume)
+-- profile.json                  # latency breakdown (setup, exec, verifier)
+-- run_meta.json                 # reward, turn count, model, ...
\end{verbatim}

The most important files for reproducibility are \texttt{events.jsonl}, which contains a strictly-ordered stream of every plan / action / observation in the run (we use it for Appendix~\ref{app:latency}), and \texttt{ctx\_state.json}, the snapshot used by the cross-session resume mechanism (Section~\ref{sec:context}). \texttt{ctx\_state.json} contains the full conversation DAG: every \texttt{ConversationNode} with its original \texttt{input}, the \texttt{compressed\_input} (if Level-1 compression was applied), and both pointer sets (\texttt{parent\_id} for the active chain, \texttt{history\_prev}/\texttt{history\_next} for the original ordering).

\paragraph{Sandbox layout.} Each invocation of \texttt{create\_sandbox} spawns an isolated process with its own filesystem rooted at \texttt{/sandbox} (the exact backing depends on the sandbox factory: local processes, Docker containers, or a managed sandbox service all expose the same interface). Files the agent uploads via \texttt{sandbox\_upload} land under \texttt{/sandbox/inputs/}; files produced by scripts go under \texttt{/sandbox/outputs/} and are pulled back with \texttt{sandbox\_download} when the agent needs them. The sandbox is destroyed at the end of the session (or earlier if the agent explicitly calls \texttt{close\_sandbox}), so no skill execution can affect host state.

\paragraph{Memory file format.} All three memory files (long-term, short-term, and per-skill) share the same plain-Markdown format: an append-only writer appends a single block of the form

\begin{verbatim}
## 2026-05-07 10:34:33 UTC
<agent-written content, one short paragraph or list>
\end{verbatim}

\noindent
to the appropriate file. Read access is line-buffered and unparsed; the agent never edits or deletes existing entries, which keeps memory append-only and makes the file safe to read from multiple sessions concurrently.

\section{Hyperparameters and Runtime Configuration}
\label{app:hyper}

Table~\ref{tab:hyperparams} lists every runtime constant used in the SkillsBench experiments. All values were held fixed across the 75-task common set; we did not perform per-task tuning. The values fall into four groups, each with a specific design intent and evaluation role.

\paragraph{Compression thresholds.} \texttt{COMPRESS\_TOKEN\_THRESHOLD} (180K) is set just below the model's hard 200K context limit, leaving a $\sim$10\% headroom so a Level-2 compression call can itself fit in context. \texttt{NODE\_COMPRESS\_TOKEN\_THRESHOLD} (15K) is the size at which a single tool output stops being amortisable across turns and starts dominating the prompt cache cost; below this threshold, leaving the original verbatim is preferable to summarising. \texttt{COMPRESS\_KEEP\_FIRST\_TURNS} and \texttt{COMPRESS\_KEEP\_LAST\_TURNS} (both 5) ensure that the task framing (the system prompt and the first few turns of grounding) and the immediate working context (the most recent five turns) are always sent verbatim; only the middle of the conversation is eligible for compression. In practice 5+5 turns are sufficient overhead even on the longest tasks we observed (max 69 turns).

\paragraph{Tool execution timeouts.} The hierarchy \texttt{TOOL\_TIMEOUT\_SECONDS} (300) $>$ \texttt{VERIFY\_COMPLETION\_TIMEOUT\_SECONDS} (120) $>$ \texttt{TERMINAL\_TIMEOUT\_SECONDS} (60) = \texttt{EXEC\_CODE\_TIMEOUT\_SECONDS} (60) reflects the expected wall-clock cost of each operation: a generic tool (e.g.\ a multi-step skill invocation) may block for several minutes, the completion checker is bounded to a single LLM call plus diagnostics, and individual shell / Python snippets are kept short to keep the ReAct loop responsive. \texttt{MODEL\_TIMEOUT\_SECONDS} (300) is a guard against API hangs; on success the actual LLM call completes in 5--30\,s. \texttt{TOOL\_TEXT\_LIMIT} (8,192 characters) is the hard cap on a single tool output before truncation, which protects the active chain from a single misbehaving tool dumping an entire log file.

\paragraph{Retry and verification.} \texttt{MAX\_RETRY} (5) is the per-call exponential-backoff budget for transient API failures (HTTP 429, 5xx). \texttt{VERIFY\_COMPLETION\_TURN\_THRESHOLD} (4) is the smallest number of turns after which the agent is allowed to call \texttt{final\_answer}; below this threshold a \texttt{verify\_completion} pre-check is forced, which prevents the agent from prematurely terminating on tasks it has barely engaged with.

\paragraph{Backbone and agents.} All four agents share the same model deployment, \texttt{gpt-5.5-2026-04-24} (paper label: GPT-5.5 (04/24/2026)). We did not set temperature, top-$p$, or other decoding overrides, so provider defaults are used throughout. The evaluated systems are \textbf{Hermes}, \textbf{Codex}, \textbf{Claude Code}, and \textbf{MUSE-Autoskill} (this work, running its own backend as described in Section~4). Claude Code's model calls are routed to \texttt{gpt-5.5-2026-04-24} through a compatibility bridge. Accuracy differences should therefore be interpreted as differences in agent prompts, tool definitions, compression policies, context handling, and skill-loading behaviour rather than differences in the model backbone. Every task is run 5 times in independent Docker containers; the SkillsBench harness controls per-task wall-clock budget.

\begin{table}[H]
\caption{All runtime constants used in the experiments. The same model-level settings were used for Hermes, Codex, Claude Code, and MUSE-Autoskill. Hermes, Codex, and Claude Code inherit only model-level constants; their tool-execution and compression behaviour is governed by their own agent systems. Tasks were graded by the SkillsBench verifier in unmodified Docker environments.}
\label{tab:hyperparams}
\centering
\small
\fitautotable{%
\begin{tabular}{lll}
\toprule
\textbf{Parameter} & \textbf{Value} & \textbf{Role} \\
\midrule
\multicolumn{3}{l}{\textit{Backbone model}} \\
\quad paper model label & GPT-5.5 (04/24/2026) & shared across all four agents \\
\quad deployment name & \texttt{gpt-5.5-2026-04-24} & shared model backend \\
\quad temperature & default & no sampling override \\[4pt]
\quad top-$p$ & default & no sampling override \\[4pt]
\multicolumn{3}{l}{\textit{Context compression (see Appendix~\ref{app:compress})}} \\
\quad \texttt{COMPRESS\_TOKEN\_THRESHOLD} & 180{,}000 & total-context trigger for Level-2 \\
\quad \texttt{NODE\_COMPRESS\_TOKEN\_THRESHOLD} & 15{,}000 & per-node trigger for Level-1 \\
\quad \texttt{COMPRESS\_KEEP\_FIRST\_TURNS} & 5 & oldest turns kept verbatim \\
\quad \texttt{COMPRESS\_KEEP\_LAST\_TURNS} & 5 & most recent turns kept verbatim \\[4pt]
\multicolumn{3}{l}{\textit{Tool execution}} \\
\quad \texttt{TOOL\_TEXT\_LIMIT} & 8{,}192 chars & per-call tool output truncation \\
\quad \texttt{TOOL\_TIMEOUT\_SECONDS} & 300 & generic tool deadline \\
\quad \texttt{TERMINAL\_TIMEOUT\_SECONDS} & 60 & shell-command deadline \\
\quad \texttt{EXEC\_CODE\_TIMEOUT\_SECONDS} & 60 & Python-snippet deadline \\
\quad \texttt{VERIFY\_COMPLETION\_TIMEOUT\_SECONDS} & 120 & deadline for the completion checker \\
\quad \texttt{MODEL\_TIMEOUT\_SECONDS} & 300 & deadline for a single LLM call \\
\quad \texttt{MAX\_RETRY} & 5 & per-call retry budget on API failures \\
\quad \texttt{VERIFY\_COMPLETION\_TURN\_THRESHOLD} & 4 & turns after which \texttt{final\_answer} requires verification \\[4pt]
\multicolumn{3}{l}{\textit{Evaluation protocol}} \\
\quad runs per task & 5 & independent Docker containers \\
\quad timeout per task & inherited from SkillsBench harness & varies by task \\
\bottomrule
\end{tabular}%
}
\end{table}

\section{SkillLearnBench Memory-System On/Off Ablation}
\label{app:memory_ablation}

We ran an additional diagnostic ablation on SkillLearnBench to isolate the effect of the MUSE-Autoskill memory system. This experiment is separate from the headline SkillLearnBench table in Section~\ref{sec:experiments}. It uses the human-skill setting together with a two-pass verifier-feedback protocol: in round~1 the agent attempts the instance normally, the verifier is run, and in round~2 a fresh environment receives the original task plus the round~1 verifier feedback. The primary verifier-feedback outcome is the round~2 result, and we also report round~1 as a pre-feedback diagnostic. The memory-system-off condition treats every instance independently. The memory-system-on condition executes instances sequentially within each task, restores the task-scoped memory state before each instance, and copies the updated memory state back after the run so that later instances can reuse accumulated observations. The round~2 prompt explicitly forbids saving oracle data, expected answers, concrete final outputs, or other instance-specific answer keys into memory.

We report all 100 runs in each condition. For each condition we report both the first attempt (round~1) and the verifier-feedback rerun (round~2). Accuracy counts verifier successes over the 100 runs in each condition; token, turn, and latency statistics summarize the corresponding runs.

\begin{table}[H]
\caption{SkillLearnBench memory-system on/off ablation under the two-pass verifier-feedback protocol. Each condition contains 100 runs; accuracy counts verifier successes, while token, turn, and latency columns summarize run-level execution cost.}
\label{tab:memory_ablation}
\centering
\scriptsize
\setlength{\tabcolsep}{3pt}
\fitautotable{%
\begin{tabular}{llrrrrrr}
\toprule
\textbf{Condition} & \textbf{Round} & \textbf{Runs} & \textbf{Successes} & \textbf{Accuracy} & \textbf{Tokens/run} & \textbf{Turns/run} & \textbf{Latency/run} \\
\midrule
Memory system off & Round 1 & 100 & 48 & 48/100 (48.0\%) & 524.4k & 12.7 & 419.1s \\
Memory system off & Round 2 & 100 & 60 & 60/100 (60.0\%) & 408.4k & 11.6 & 338.6s \\
\rowcolor{autoskillblue}
Memory system on & Round 1 & 100 & 64 & 64/100 (64.0\%) & 467.3k & 11.4 & 347.7s \\
\rowcolor{autoskillblue}
Memory system on & Round 2 & 100 & 74 & \textbf{74/100 (74.0\%)} & 423.9k & 11.7 & 295.4s \\
\bottomrule
\end{tabular}%
}
\end{table}

In this experiment, the memory system improves the round~2 verifier-feedback accuracy by $+14.0$ percentage points (60/100 $\rightarrow$ 74/100). The same pattern is already visible before feedback: round~1 accuracy rises from 48/100 (48.0\%) to 64/100 (64.0\%). The memory system also reduces round~1 execution cost (524.4k $\rightarrow$ 467.3k tokens/run; 12.7 $\rightarrow$ 11.4 turns/run; 419.1s $\rightarrow$ 347.7s). In round~2, tokens are slightly higher with the memory system (408.4k $\rightarrow$ 423.9k), turns are essentially unchanged (11.6 $\rightarrow$ 11.7), and latency is lower (338.6s $\rightarrow$ 295.4s). Overall, the result suggests that the memory system improves reliability on repeated SkillLearnBench tasks without increasing the number of reasoning turns.

\section{Compression Algorithm}
\label{app:compress}

Context compression is invoked by \texttt{maybe\_compress\_history(ctx, model)} at the start of every ReAct turn, immediately after the agent's response is appended to the conversation and before the next LLM call is issued. The function returns silently when the active chain is under budget (which is the common case at the start of a run) and only triggers an LLM-summarisation call when the total token estimate crosses \texttt{COMPRESS\_TOKEN\_THRESHOLD}. We implement two levels of progressively more aggressive compression; in our SkillsBench runs Level-1 is sufficient for the vast majority of contexts that exceed the budget and Level-2 fires only on the longest-running tasks (turn count $>$50). Both levels operate exclusively on the active chain (the linked list reachable via \texttt{parent\_id}); the immutable \texttt{history\_prev}/\texttt{history\_next} pointers are never rewritten, so any prior state can still be reconstructed for cross-session resume or for post-hoc trajectory analysis. The high-level control flow is:

{\small
\begin{verbatim}
def maybe_compress_history(ctx, model):
    chain        = walk(parent_id from tip to root)
    total_tokens = sum(estimate_tokens(node) for node in chain)
    if total_tokens <= COMPRESS_TOKEN_THRESHOLD:
        return                                  # under budget; nothing to do

    # ---- Level 1: per-node, in-place summary on oversized nodes ----
    # never touch the first K=5 or last K=5 turns
    middle = chain[KEEP_FIRST : -KEEP_LAST]
    for node in middle:
        if estimate_tokens(node) > NODE_COMPRESS_TOKEN_THRESHOLD:
            summary = model.summarize(node.input + node.model_output)
            node.compressed_input   = summary
            node.is_node_compressed = True      # reads return summary

    if recompute_total(chain) <= COMPRESS_TOKEN_THRESHOLD:
        return                                  # Level 1 was enough

    # ---- Level 2: collapse the middle span into one summary node ----
    span    = chain[KEEP_FIRST : -KEEP_LAST]
    summary = model.summarize(concat(span))
    sNode   = new ConversationNode(
        is_summary       = True,
        parent_id        = chain[KEEP_FIRST - 1].node_id,
        compressed_input = summary,
    )
    chain[-KEEP_LAST].parent_id = sNode.node_id  # rewire chain
\end{verbatim}
}

\paragraph{Cost.} Compression itself costs LLM calls: Level-1 issues at most one summarisation call per oversized node, Level-2 issues exactly one summarisation call per trigger. Because the threshold is much larger than a typical tool output, the amortised cost is small (one extra LLM call every $\sim$10--20 ReAct turns on the long-running tasks we observed). The summary calls use the same backbone model (GPT-5.5 (04/24/2026)) at provider defaults; we did not separately tune them.

\paragraph{Audit trail.} The original \texttt{node.input} field is never mutated. Reads through the active-chain reader return \texttt{compressed\_input} when \texttt{is\_node\_compressed} is True, so the active chain shrinks; reads through the full-history reader ignore the \texttt{compressed\_input} field and walk the immutable history pointers, so the original ordering is recoverable. Level-2's synthetic summary node has \texttt{is\_summary=True} and, by construction, \emph{no} history pointers, so the full-history reader skips it and recovers exactly the original sequence of turns. This means any compressed run can be ``replayed'' for analysis without re-running the agent.

\paragraph{Why not just truncate?} A simpler alternative (drop the oldest middle turns when the budget is exceeded) would lose information silently. Our preliminary experiments showed that on multi-step tasks the agent revisits early-context facts roughly 30--40\% of the time (e.g.\ to recheck an input filename or recall a parsing-format detail); truncation forced wasteful re-discovery. Summarisation preserves these facts at $\sim$1/10 the token cost, which is the regime where the LLM-call overhead pays for itself.

\section{Detailed Token Breakdown}
\label{app:tokens}

Token accounting is diagnostic rather than a headline comparison because fresh/cache token traces are complete for MUSE-Autoskill and mostly complete for Codex, while Hermes and Claude Code expose total-token metadata without the same normalized fresh/cache decomposition in the current logs. Table~\ref{tab:token_breakdown} reports medians over available runs from the 75-task common set where that decomposition is parseable, splitting prompt tokens into the \emph{fresh} component and the \emph{cached} component, and reporting output and reasoning tokens separately for each condition.

Two patterns are worth flagging. First, both MUSE-Autoskill and Codex show substantial prompt-cache use: median cached input is larger than median fresh input in every reported condition. Second, human skills increase MUSE-Autoskill's median token footprint, primarily through additional skill-catalog and skill-body context, while Codex changes only modestly. Since this detailed breakdown excludes total-only Hermes and Claude Code component rows, we use latency rather than token cost for the complete four-agent efficiency comparison in Appendix~\ref{app:latency}.

\begin{table}[H]
\caption{Per-task token usage diagnostic on the 75-task common set. Columns are medians computed independently over runs with parseable token traces, so ``Total'' need not equal the sum of the displayed median components. ``Reasoning'' is counted within ``output'' by the model API.}
\label{tab:token_breakdown}
\centering
\fitautotable{%
\begin{tabular}{llrrrrrr}
\toprule
\textbf{Agent} & \textbf{Condition} & \textbf{Fresh in} & \textbf{Cached in} & \textbf{Output} & \textbf{Reasoning} & \textbf{Total} & \textbf{$n$} \\
\midrule
Codex      & without skills & 133,211 & 210,880 & 7,676 & 2,246 & 342,220 & 359 \\
Codex      & human skills   & 146,031 & 199,680 & 7,007 & 1,987 & 360,683 & 361 \\
\rowcolor{autoskillblue}
MUSE-Autoskill       & without skills & 222,155 & 322,688 & 13,314 & 4,986 & 579,499 & 375 \\
\rowcolor{autoskillblue}
MUSE-Autoskill       & human skills   & 266,030 & 359,424 & 13,059 & 4,472 & 638,000 & 375 \\
\bottomrule
\end{tabular}
}
\end{table}

\section{Per-Domain Accuracy with Standard Deviation}
\label{app:perdomain}

Table~\ref{tab:perdomain_std} reports the per-domain mean accuracy together with task-level standard deviation, for each of the four agents and both skill conditions. Standard deviation is computed across the per-task means (each task's mean is itself averaged over 5 runs); a high $\sigma$ signals that the agent's accuracy varies considerably across tasks in that domain (some tasks land near 100\%, others near 0\%) and does not directly reflect run-to-run noise.

Skills improve all four domains for all four agents in aggregate. The largest domain lift is in Ops \& Planning for Codex, Claude Code, and MUSE-Autoskill, while Hermes gains most in Science \& Engineering. MUSE-Autoskill leads the human-skill column in Science \& Engineering, Document Processing, and Ops \& Planning; Claude Code leads Data Analysis. High standard deviations remain common because each domain mixes near-solved tasks with tasks that no agent solves reliably.

\begin{table}[H]
\caption{Per-domain accuracy (\%) with task-level standard deviation on the 75-task common set. Lift is the difference of means (w/ human skills $-$ w/o skills).}
\label{tab:perdomain_std}
\centering
\setlength{\tabcolsep}{4pt}
\fitautotable{%
\begin{tabular}{llcccc}
\toprule
\textbf{Agent} & \textbf{Domain} & \textbf{$n$ tasks} & \textbf{w/o skills} & \textbf{w/ human skills} & \textbf{Lift} \\
\midrule
Hermes & Science \& Engineering & 18 & 40.61 $\pm$ 45.29 & 58.89 $\pm$ 39.71 & +18.28 \\
Hermes & Data Analysis & 18 & 36.34 $\pm$ 43.30 & 40.60 $\pm$ 40.09 & +4.26 \\
Hermes & Document Processing & 14 & 54.29 $\pm$ 37.36 & 58.57 $\pm$ 41.03 & +4.29 \\
Hermes & Ops \& Planning & 25 & 25.92 $\pm$ 36.87 & 39.64 $\pm$ 36.31 & +13.72 \\
\midrule
Codex & Science \& Engineering & 18 & 54.52 $\pm$ 43.14 & 67.12 $\pm$ 40.28 & +12.60 \\
Codex & Data Analysis & 18 & 38.31 $\pm$ 41.38 & 50.18 $\pm$ 44.48 & +11.87 \\
Codex & Document Processing & 14 & 68.57 $\pm$ 33.56 & 72.86 $\pm$ 32.61 & +4.29 \\
Codex & Ops \& Planning & 25 & 29.16 $\pm$ 34.41 & 47.48 $\pm$ 41.00 & +18.32 \\
\midrule
Claude Code & Science \& Engineering & 18 & 52.33 $\pm$ 40.07 & 63.88 $\pm$ 38.72 & +11.55 \\
Claude Code & Data Analysis & 18 & 38.05 $\pm$ 39.63 & 52.59 $\pm$ 44.05 & +14.54 \\
Claude Code & Document Processing & 14 & 62.86 $\pm$ 39.90 & 64.29 $\pm$ 37.17 & +1.43 \\
Claude Code & Ops \& Planning & 25 & 27.00 $\pm$ 36.78 & 48.60 $\pm$ 41.82 & +21.60 \\
\midrule
\rowcolor{autoskillblue}
MUSE-Autoskill & Science \& Engineering & 18 & 54.57 $\pm$ 40.93 & 67.97 $\pm$ 40.99 & +13.40 \\
\rowcolor{autoskillblue}
MUSE-Autoskill & Data Analysis & 18 & 39.49 $\pm$ 44.36 & 51.48 $\pm$ 47.02 & +11.99 \\
\rowcolor{autoskillblue}
MUSE-Autoskill & Document Processing & 14 & 67.14 $\pm$ 35.14 & 74.29 $\pm$ 34.99 & +7.14 \\
\rowcolor{autoskillblue}
MUSE-Autoskill & Ops \& Planning & 25 & 35.52 $\pm$ 40.80 & 51.40 $\pm$ 39.78 & +15.88 \\
\bottomrule
\end{tabular}
}
\end{table}

\section{Latency and Turn-Count Distribution}
\label{app:latency}

Table~\ref{tab:latency_dist} reports the percentile distribution of per-task latency (agent execution time in seconds, excluding the SkillsBench verifier) and ReAct turn counts. Both are computed across 75 tasks $\times$ 5 runs $=$ 375 runs per (agent, condition) cell. \textbf{Hermes} is the fastest runtime by median latency, followed by Claude Code, MUSE-Autoskill, and Codex. Claude Code has a longer high-percentile tail than Hermes despite a comparable median, while Codex remains the slowest median runtime. MUSE-Autoskill runs deeper loops than Hermes and Claude Code, but its tail is shorter than Codex's.

A second observation is that human skills reduce median latency for every agent while improving accuracy: Hermes drops from 354.0\,s to 327.3\,s, Codex from 1013.6\,s to 869.5\,s, Claude Code from 347.3\,s to 291.4\,s, and MUSE-Autoskill from 747.6\,s to 730.8\,s. Turn counts do not uniformly decrease, so the complete efficiency claim is about wall-clock latency rather than fewer ReAct steps.

\begin{table}[H]
\caption{Distribution of per-task agent latency (seconds) and ReAct turn counts, by agent and skill condition on the 75-task common set. ``p10'' / ``p25'' / ``p75'' / ``p90'' are the 10th, 25th, 75th, and 90th percentile. ``max'' for turns is the absolute maximum observed across all runs in that cell.}
\label{tab:latency_dist}
\centering
\small
\setlength{\tabcolsep}{3.5pt}
\fitautotable{%
\begin{tabular}{llcccccccccc}
\toprule
\textbf{Agent} & \textbf{Cond.} & \multicolumn{5}{c}{\textbf{Latency (s)}} & \multicolumn{4}{c}{\textbf{Turns}} & \textbf{$n$} \\
\cmidrule(lr){3-7}\cmidrule(lr){8-11}
& & p10 & p25 & median & p75 & p90 & p25 & median & p75 & max & \\
\midrule
Hermes & w/o & 92.5 & 183.2 & 354.0 & 668.7 & 1084.2 & 9 & 14 & 19 & 73 & 375 \\
Hermes & w/ & 102.9 & 166.8 & 327.3 & 545.8 & 1063.8 & 9 & 13 & 18 & 61 & 375 \\
\midrule
Codex & w/o & 350.8 & 537.3 & 1013.6 & 1772.8 & 2926.1 & 8 & 12 & 19 & 57 & 375 \\
Codex & w/ & 275.8 & 457.1 & 869.5 & 1641.0 & 2596.0 & 8 & 14 & 22 & 47 & 375 \\
\midrule
Claude Code & w/o & 111.4 & 168.1 & 347.3 & 884.2 & 1868.2 & 12 & 19 & 28 & 227 & 375 \\
Claude Code & w/ & 91.7 & 148.4 & 291.4 & 850.8 & 1929.1 & 14 & 22 & 32 & 109 & 375 \\
\midrule
\rowcolor{autoskillblue}
MUSE-Autoskill & w/o & 245.7 & 454.4 & 747.6 & 1172.3 & 1836.2 & 15 & 20 & 26 & 72 & 375 \\
\rowcolor{autoskillblue}
MUSE-Autoskill & w/ & 241.3 & 413.5 & 730.8 & 1166.3 & 1676.6 & 15 & 20 & 26 & 82 & 375 \\
\bottomrule
\end{tabular}%
}
\end{table}

\section{Self-Created Skill Stability}
\label{app:stability}

Table~\ref{tab:self_created_stability} compares run-to-run stability for self-created skills on each agent's own covered subset. For each task, we compute the mean reward and standard deviation over the canonical five independent runs, then average these task-level statistics within the covered subset. Lower standard deviation and lower mean absolute deviation indicate that repeated runs produce more consistent rewards, regardless of whether the mean is 3/5, 4/5, or 5/5. MUSE-Autoskill has both the highest covered-task reward and the lowest run-to-run dispersion: its average per-task standard deviation is 0.109, compared with 0.124 for Codex and 0.182 for Claude Code. It also has the fewest non-constant tasks (13/47) and the most low-variance tasks (35/47).

\begin{table}[H]
\caption{Run-to-run stability of self-created skills on each agent's covered subset. Rewards are computed over the canonical five independent runs per task. ``Non-constant'' counts tasks whose five rewards are not identical. ``Low-var.'' counts tasks with per-task reward standard deviation at most 0.1.}
\label{tab:self_created_stability}
\centering
\small
\setlength{\tabcolsep}{5pt}
\fitautotable{%
\begin{tabular}{lrrrrrr}
\toprule
\textbf{Agent} & \textbf{Covered} & \textbf{Avg. Reward} & \textbf{Avg. Std.} & \textbf{Avg. MAD} & \textbf{Non-constant} & \textbf{Low-var.} \\
\midrule
Codex & 47 & 75.83\% & 0.124 & 0.110 & 17/47 & 32/47 \\
Claude Code & 44 & 75.45\% & 0.182 & 0.160 & 21/44 & 26/44 \\
\rowcolor{autoskillblue} MUSE-Autoskill (Ours) & 47 & \textbf{85.24\%} & \textbf{0.109} & \textbf{0.096} & \textbf{13/47} & \textbf{35/47} \\
\bottomrule
\end{tabular}%
}
\end{table}

This stability pattern suggests that MUSE-Autoskill-created skills often encode a more executable procedure than general skill descriptions. Human- or agent-authored skills can provide useful domain knowledge while still leaving run-specific choices to the agent. In contrast, a skill distilled from a successful trajectory tends to preserve concrete procedural details, such as command sequences, file paths, output schemas, validation checks, and task-specific failure modes. These details narrow the action space during reuse, making repeated executions more likely to follow the same successful path. The benefit is not universal: trajectory-derived skills can also overfit to brittle source-run assumptions, as illustrated by the \texttt{hvac-control} regression case study in Section~\ref{sec:skill_gen} on held-out reruns.

\section{Skill-Generation Failures: The 28 MUSE-Uncovered Tasks}
\label{app:failures}

MUSE-Autoskill produced no usable self-created skill for 28 of the 75 tasks under the strict 75-task protocol. These tasks contribute 0\% to the all-task average in Table~\ref{tab:skill_gen}. Their distribution characterises the current limits of inference-time skill synthesis.

The uncovered set is concentrated in Ops \& Planning and Data Analysis, with smaller clusters in Science \& Engineering and Document Processing. This supports the bottleneck analysis in Section~\ref{sec:skill_gen}: self-created skills are strong when a reusable source trajectory exists, but coverage still depends on Phase~1 exploration finding such a trajectory. A natural next direction is to extract \emph{partial} skills from failed trajectories, capturing diagnostic moves that worked even when the run ultimately ended at reward 0.

\end{document}